\documentclass{article}

\usepackage{arxiv}

\usepackage[utf8]{inputenc} % allow utf-8 input
\usepackage[T1]{fontenc}    % use 8-bit T1 fonts
\usepackage{hyperref}       % hyperlinks
\usepackage{url}            % simple URL typesetting
\usepackage{booktabs}       % professional-quality tables
\usepackage{amsfonts}       % blackboard math symbols
\usepackage{nicefrac}       % compact symbols for 1/2, etc.
\usepackage{microtype}      % microtypography
\usepackage{xcolor}         % colors

\usepackage[american]{babel}
\usepackage{amsmath}
\usepackage{amssymb}
\usepackage{amsthm}
\usepackage{tikz}
\usetikzlibrary{arrows,cd,decorations.markings,shapes.geometric,shapes}
\usepackage{mathtools}
\usepackage{ebproof}

\usepackage{tikz}
\usetikzlibrary{arrows,cd,decorations.markings,shapes.geometric,shapes}

\newcommand\cofib\rightarrowtail

\newcommand\mdel[1]{}

\usepackage{txfonts}
\usepackage{pxfonts}

\makeatletter
\newcommand{\xdashrightarrow}[2][]{\ext@arrow 0359\rightarrowfill@@{#1}{#2}}
\newcommand*{\doublerightarrow}[2]{\mathrel{
  \settowidth{\@tempdima}{$\scriptstyle#1$}
  \settowidth{\@tempdimb}{$\scriptstyle#2$}
  \ifdim\@tempdimb>\@tempdima \@tempdima=\@tempdimb\fi
  \mathop{\vcenter{
    \offinterlineskip\ialign{\hbox to\dimexpr\@tempdima+1em{##}\cr
    \rightarrowfill\cr\noalign{\kern.5ex}
    \rightarrowfill\cr}}}\limits^{\!#1}_{\!#2}}}
\newcommand*{\triplerightarrow}[1]{\mathrel{
  \settowidth{\@tempdima}{$\scriptstyle#1$}
  \mathop{\vcenter{
    \offinterlineskip\ialign{\hbox to\dimexpr\@tempdima+1em{##}\cr
    \rightarrowfill\cr\noalign{\kern.5ex}
    \rightarrowfill\cr\noalign{\kern.5ex}
    \rightarrowfill\cr}}}\limits^{\!#1}}}
\makeatother

\makeatletter
\newcommand{\twoarrows}[3][0.2ex]{%
  \mathrel{\mathpalette\twoarrows@{{#1}{#2}{#3}}}%
}
\newcommand{\twoarrows@}[2]{\twoarrows@@#1#2}
\newcommand{\twoarrows@@}[4]{%
  \vcenter{\offinterlineskip\m@th
    \ialign{\hfil##\hfil\cr
      $#1#3$\cr
      \noalign{\vskip#2}
      $#1#4$\cr
    }%
  }%
}
\makeatother

\newcommand{\beq}{\begin{equation}}
\newcommand{\eeq}{\end{equation}}

\newtheorem{theorem}{Theorem}
\newtheorem{definition}{Definition}
\newtheorem{lemma}{Lemma}
\newtheorem{example}{Example}

\usepackage[boxruled]{algorithm2e}
\usepackage{algorithmic}

\usepackage{multirow}
\usepackage{booktabs}
\usepackage{balance}
\usepackage{subfig}

\DeclareFontFamily{U}{dmjhira}{}
\DeclareFontShape{U}{dmjhira}{m}{n}{ <-> dmjhira }{}

\usepackage{natbib} % has a nice set of citation styles and commands
\usepackage{mathtools} % amsmath with fixes and additions
\usepackage{booktabs} % commands to create good-looking tables
\usepackage{tikz} % nice language for creating drawings and diagrams

\title{Topos Theory for Generative AI and LLMs\thanks{Draft under submission.} }

\author{ Sridhar Mahadevan \\
	Adobe Research and University of Massachusetts, Amherst\\
	\texttt{smahadev@adobe.com, mahadeva@umass.edu}
}

\hypersetup{
pdftitle={A template for the arxiv style},
pdfsubject={q-bio.NC, q-bio.QM},
pdfauthor={David S.~Hippocampus, Elias D.~Striatum},
pdfkeywords={First keyword, Second keyword, More},
}

\begin{document}
\maketitle

\begin{abstract}
We propose the design of novel generative AI architectures (GAIA)  using {\em topos theory}, a type of category that is ``set-like": a topos has all (co)limits, is Cartesian closed, and has a subobject classifier. Previous theoretical results on the Transformer model have shown that it is a universal sequence-to-sequence function approximator, and dense in the space of all continuous functions on $\mathbb{R}^{d \times n}$ with compact support. Building on this theoretical result, we explore novel architectures for LLMs that exploit the property that the category of LLMs, viewed as functions, forms a topos.  Previous studies of large language models (LLMs) have focused on daisy-chained linear architectures or mixture-of-experts. In this paper, we use the theory of categories and functors  to construct much richer LLM architectures based on new types of compositional structures. In particular, these new compositional structures are derived from universal properties of LLM categories, and include {\em pullback}, {\em pushout}, {\em (co) equalizers}, exponential compositions, and subobject classifiers.  We theoretically validate these new compositional structures by showing that the category of LLMs is (co)complete, meaning that all diagrams have solutions in the form of (co)limits.  Building on this completeness result, we then show that the category of LLMs forms a topos, a ``set-like" category, which requires showing the existence of exponential objects as well as subobject classifiers. We use a functorial characterization of backpropagation to define the implementation of an LLM topos architecture. 
\end{abstract}

% keywords can be removed

\newpage 

\tableofcontents

\newpage

\section{Introduction}

\begin{figure}[t]
    \centering
    \includegraphics[width=0.9\linewidth]{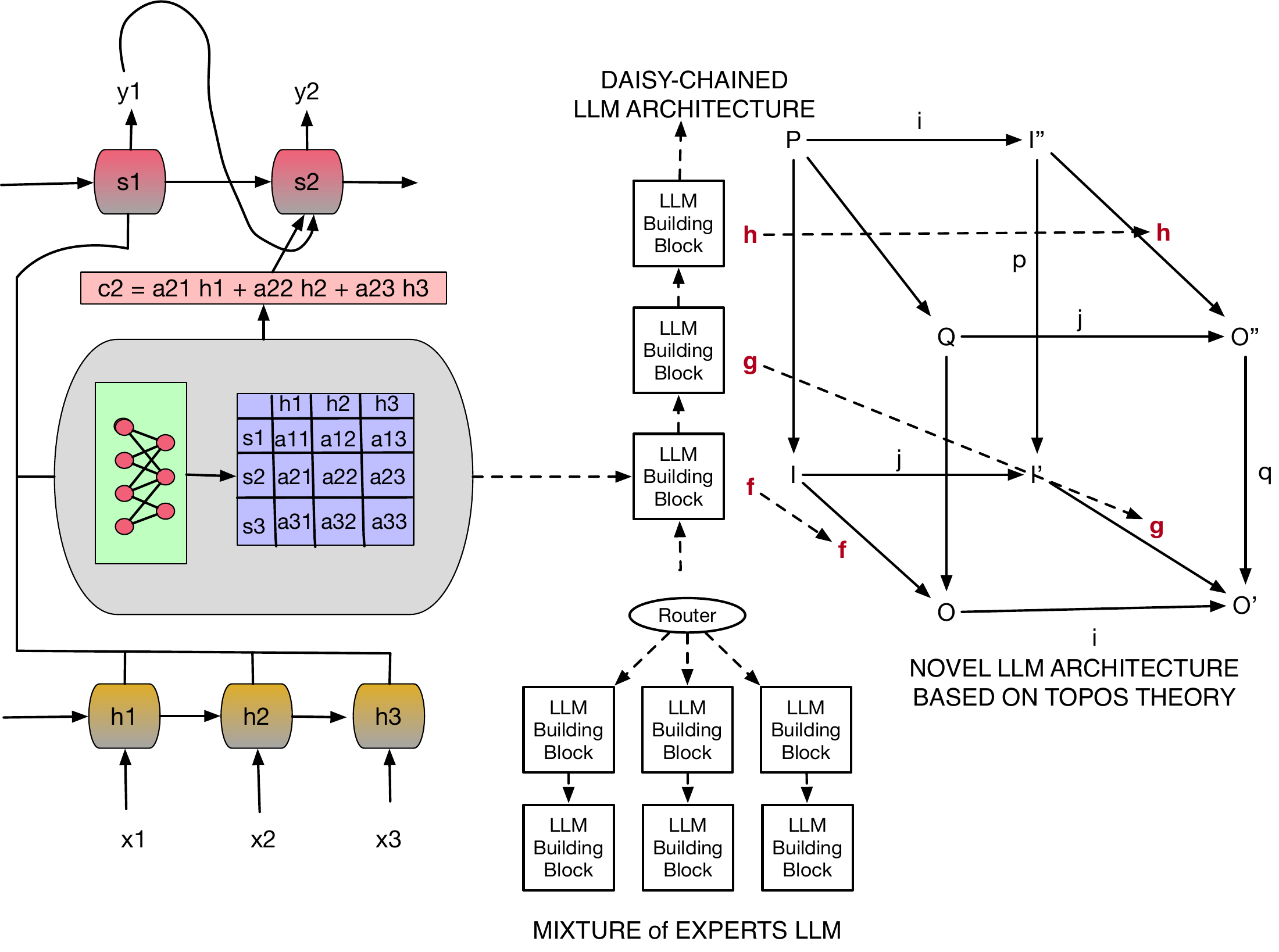}
    \caption{LLMs are typically a daisy-chained sequence of primitive building blocks \citep{chaudhari2021attentivesurveyattentionmodels}, or mixture-of-experts (MOE) \citep{deepseekai2025deepseekr1incentivizingreasoningcapability,wang2024mixtureofagentsenhanceslargelanguage}. We use topos theory to create novel LLM architectures, e.g., a ``cube".}
    \label{fig:llm-arch}
\end{figure}

We propose the use of {\em topos theory} to design novel categorical architectures for generative AI (GAIA), extending our previous work \citep{mahadevan2024gaiacategoricalfoundationsgenerative}, which did not use any concepts from topos theory. Toposes \citep{maclane:sheaves,bell,goldblatt:topos,Johnstone:topostheory} are categories that resemble sets. Just as in sets, you are allowed operations like set intersection and union, in a topos, you can construct limits and colimits, universal constructions \citep{riehl2017category}. A set can always be decomposed into subsets: the generalization of subsets in a topos is given by a {\em subobject classifier}. Finally, the set of all functions $f: A \rightarrow B$ from a set $A$ to another set $B$ is itself a set, or an exponential object. A topos is required to have an exponential object for any two objects $c, d \in {\cal C}$. The primary goal of this paper is to illustrate how these ideas can be applied to generative AI to design novel architectures. We will use the Transformer model \citep{DBLP:conf/nips/VaswaniSPUJGKP17,attention-turing-complete,chaudhari2021attentivesurveyattentionmodels,merrill-etal-2022-saturated,DBLP:conf/iclr/YunBRRK20} as our primary example, although the basic framework applies to other generative AI models, such as structured state space sequence models \citep{DBLP:conf/iclr/GuGR22} (see Figure~\ref{fig:llm-arch}). 

In recent years, large language models (LLMs) using the Transformer model \citep{DBLP:conf/nips/VaswaniSPUJGKP17} have enabled building large foundation models \citep{fm}, and has led to the prospect of artificial general intelligence (AGI) \citep{agi-dm}.  Theoretical studies have shown finite precision LLMs cannot recognize Dyck languages or compute the parity function  \citep{DBLP:journals/tacl/Hahn20,merrill2022saturatedtransformersconstantdepththreshold}, and on the other hand infinite precision LLMs are Turing complete \citep{attention-turing-complete} and  that they are a universal function approximator over sequences \citep{DBLP:conf/iclr/YunBRRK20}. A number of studies have shown that LLMs exhibit characteristic failures in reasoning \citep{faithandfate}, and efforts to develop a deeper theoretical understanding of non-compositional behavior of LLMs are ongoing \citep{llmcompositionality}. 

Most previous work have been restricted to empirical or theoretical study of a simple daisy-chained sequential LLM architectures. There also has been growing interest in other architectures, most commonly the well-studied {\em mixture of experts} (MOE) architecture \citep{wang2024mixtureofagentsenhanceslargelanguage} used in DeepSeek-R1 \citep{deepseekai2025deepseekr1incentivizingreasoningcapability}. Here, a {\em router} guides input tokens to some ``expert" LLM.  In contrast, our paper uses the rigorous framework of category theory \citep{maclane:71,riehl2017category} to introduce a novel class of LLM architectures that derive from {\em universal constructions}. Since \citet{DBLP:conf/iclr/YunBRRK20} showed that LLMs are a universal function approximator on sequences, for simplicity, we can model each LLM building block as a function $f,g,h$ as shown in the center of Figure~\ref{fig:llm-arch}. We can use universal constructions in category theory to explore a much richer class, such as the ``cube" architecture. In the cube LLM architecture, we are given the basic LLM building blocks, shown as ``arrows" $f$, $g$, and $h$ mapping some input sequence $I$ or $I'$ or $I''$, respectively, to some output sequence $O$, $O'$, and $O"$, respectively. A number of categorical studies of deep learning have been published recently \citep{DBLP:conf/lics/FongST19,gavranović2024positioncategoricaldeeplearning,mahadevan2024gaiacategoricalfoundationsgenerative}, but these have not used topos theory as a way to generate novel  architectures. 

\section{LLM as a Category}
\label{llmcat-defns}

To explain our approach, we need to introduce some basics of category theory \citep{maclane:71}, as well as topos theory \citep{maclane:sheaves}. A primer for readers not familiar with the necessary mathematical background is given in Section~\ref{introcat}.  We use examples from LLMs to illustrate the abstract theory. A {\em category} ${\cal C}$ \citep{maclane:71} is simply a collection of abstract {\em objects}, e.g, $c, d \in {\cal C}$, along with a collection of abstract {\em arrows} ${\cal C}(c,d)$ between each pair of objects. As we will see, what constitutes an ``object" or an ``arrow" is very flexible, leading to a highly expressive representational capacity.  To illustrate how to form a category based on LLMs, let us review the basic design of Transformer models \citep{DBLP:conf/nips/VaswaniSPUJGKP17,chaudhari2021attentivesurveyattentionmodels}. 
\begin{definition}
    A {\bf Transformer} block is a sequence-to-sequence function mapping $\mathbb{R}^{d \times n} \rightarrow \mathbb{R}^{d \times n}$. There are generally two layers: a {\em self-attention} layer and a token-wise feedforward layer. We assume tokens are embedded in a space of dimension $d$. Specifically, we model the inputs $X \in \mathbb{R}^{d \times n}$ to a Transformer block as $n$-length sequences of tokens in $d$ dimensions, where each block computes the following function defined as $t^{h,m,r}: \mathbb{R}^{d \times n}: \mathbb{R}^{d \times n}$: 

\begin{eqnarray*}
    \mbox{Attn}(X) &=& X + \sum_{i=1}^h W^i_O W^i_V X \cdot \sigma[W^i_K X)^T W^i_Q X] \\
    \mbox{FF}(X) &=& \mbox{Attn}(X) + W_2 \cdot \mbox{ReLU}(W_1 \cdot \mbox{Attn}(X) + b_1 {\bf 1}^T_n, 
\end{eqnarray*}

where $W^i_O \in \mathbb{R}^{d \times n}$, $W^i_K, W^i_Q, W^i_Q \in \mathbb{R}^{d \times n}$, $W_2 \in \mathbb{R}^{d \times r}$, $W_1 \in \mathbb{R}^{r \times d}$, and $b_1 \in \mathbb{R}^r$. The output of a Transformer block is $FF(X)$. Following convention, the number of ``heads" is $h$, and each ``head"  size $m$
are the principal parameters of the attention layer, and the size of the ``hidden" feed-forward layer is $r$. 
\end{definition}
Transformer models take as input objects $X \in \mathbb{R}^{d \times n}$ representing $n$-length sequences of tokens in $d$ dimensions, and act as morphisms that represent permutation equivariant functions $f: \mathbb{R}^{d \times n} \rightarrow \mathbb{R}^{d \times n}$ such that $f(XP) = f(X)P$ for any permutation matrix $P$.  \citet{DBLP:conf/iclr/YunBRRK20} show that the actual function computed by the Transformer model defined above is a permutation equivariant mapping.  We can define permutation equivariance through the following commutative diagram: 
% https://q.uiver.app/#q=WzAsNixbMCwwLCJYIl0sWzIsMCwiWSJdLFs0LDAsIloiXSxbMCwyLCJYUCJdLFsyLDIsIllQIl0sWzQsMiwiWlAiXSxbMCwxLCJmIl0sWzEsNCwiUCJdLFswLDMsIlAiLDJdLFszLDQsImYiLDJdLFsxLDIsImciXSxbNCw1LCJnIiwyXSxbMiw1LCJQIl1d
\[\begin{tikzcd}
	X && Y && Z \\
	\\
	XP && YP && ZP
	\arrow["f", from=1-1, to=1-3]
	\arrow["P", from=1-3, to=3-3]
	\arrow["P"', from=1-1, to=3-1]
	\arrow["f"', from=3-1, to=3-3]
	\arrow["g", from=1-3, to=1-5]
	\arrow["g"', from=3-3, to=3-5]
	\arrow["P", from=1-5, to=3-5]
\end{tikzcd}\]
In the above commutative diagram, vertices $X, Y$ etc. are objects, and arrows $f,g$ etc. are morphisms that define the action of a Transformer block. Here, $X \in \mathbb{R}^{d \times n}$ is a $n$-length sequence of tokens of dimensionality $d$. $P$ is a permutation matrix. The function $f$ computed by a Transformer block is such that $f(XP) = f(X) P$. Permutation equivariance makes the diagram above commute:  we can compute $YP$ by first computing $Y = f(X)$, and then applying the permutation matrix $P$, or first permuting $X$ to obtain $XP$ and the computing $YP = f(XP)$.

There are many ways to define categories of Transformer models. \citet{bradley:enriched-yoneda-llms} for instance define an LLM category based on the next-token distribution probabilities, where objects are fragments of sentences $x = \text{I am flying}$, and $y = \text{I am flying to Singapore}$ is a possible completion. Then the arrow ${\cal C}(x,y) = P(y | x)$, the conditional probability of completing $x$ by $y$. In our paper, we focus on the representational capacity of Transformers as universal sequence-to-sequence function approximators, following \citep{DBLP:conf/iclr/YunBRRK20}. Thus, we will choose to define the category of Transformers ${\cal C}^\rightarrow_T$ in the following way. 
\begin{definition} \label{transformercat}
    The category ${\cal C}^\rightarrow_T$ of Transformer models is defined as the following category: 
    \begin{itemize}
        \item The objects $\mbox{Obj}({\cal C}^\rightarrow_T)$ are defined as functions $f,g$ mapping between token sequences in  $\mathbb{R}^{d \times n}$ (as functions are treated as ``objects" here, this type of category is sometimes referred to as an ``arrow" category \citep{maclane:71}, which we highlight by using the $\rightarrow$ in ${\cal C}^\rightarrow_T$ ). 

        \item The arrows of the category ${\cal C}^\rightarrow_T$ are defined as commutative diagrams of the type shown above (which compose horizontally by adding boxes). 
 
 \begin{center}
 \label{gdcarrow}
%https://q.uiver.app/#q=WzAsNCxbMCwwLCJVIl0sWzAsMiwiViJdLFsyLDAsIlUnIl0sWzIsMiwiViciXSxbMCwxLCJmIl0sWzAsMiwiaCIsMl0sWzIsMywiZiciLDJdLFsxLDMsImciXV0=
\begin{tikzcd}
	I && {I'} \\
	\\
	O && {O'}
	\arrow["h"', from=1-1, to=1-3]
	\arrow["f", from=1-1, to=3-1]
	\arrow["{f'}"', from=1-3, to=3-3]
	\arrow["g", from=3-1, to=3-3]
\end{tikzcd}
 \end{center}
 In the above diagram, the functions $f,f'$ are computed by two Transformer models, and $g,h$ are mappings between token sequences in $\mathbb{R}^{d \times n}$ that make the diagram commute.
    \end{itemize}
\end{definition}
\section{Dense Functors: LLMs as Universal Function Approximators} 
A large number of recent studies have explored the theoretical properties of LLMs. \citet{attention-turing-complete} show that attention is Turing complete, assuming that the architecture can compute with arbitrary real numbers. In contrast, \citet{merrill-etal-2022-saturated} show that given finite precision bounded by the logarithm of the number of input tokens, Transformers can recognize languages only within a fairly limited circuit complexity class (e.g., AC0 or TC0). \citep{transformer-bounds} derive bounds on Transformers by counting quantifiers in first-order logic, building on the connections between logic and computational complexity.  

In our paper, for simplicity, we focus on modeling an LLM as a sequence-to-sequence function, as defined above. \citet{DBLP:conf/iclr/YunBRRK20} showed that LLMs can approximate any continuous function from sequences to sequences. Our design of novel category theoretic LLM architectures will require building on results showing LLMs are universal sequence-to-sequence function approximators. First, let us introduce some notation from \citep{DBLP:conf/iclr/YunBRRK20}. 
\begin{definition}\citep{DBLP:conf/iclr/YunBRRK20}
The function class ${\cal F}_{PE}$ consists of all permutation equivariant functions with compact support that maps $\mathbb{R}^{d \times n}$ to  $\mathbb{R}^{d \times n}$ , where continuity is defined with respect to an entry-wise $l^p$ norm, $1 \leq p < \infty$. 
\end{definition}
\begin{definition}
Define the building block of Transformers as $t^{h,m,r}$, which denotes a Transformer block defined by an attention layer with $h$ heads of size $m$,  and  each, with a feedforward layer of $r$ hidden nodes.    Then, the function class is defined as 
\[ {\cal T}^{h,m,r} = \{ g: \mathbb{R}^{d \times n} \rightarrow \mathbb{R}^{d \times n} | g \ \mbox{is a composition of} \ t^{h, m, r } \} \] 
\end{definition}
\begin{theorem}\citep{DBLP:conf/iclr/YunBRRK20}
Let $1 \leq p < \infty$ and $\epsilon > 0$, then for any function $f \in {\cal F}_{PE}$, there exists a Transformer network $g \in {\cal T}^{2,1,4}$ such that $d_p(f,g) < \epsilon$, where 
\[ d_p(f_1, f_2) = \left( \int \| f_1({\bf X}) - f_2({\bf X})\|^p_p d {\bf X} \right)^{\frac{1}{p}}\]
\end{theorem}
To overcome the restriction to permutation-equivariant functions, it is common in Transformer implementations to include a Relative Position Embedding function (RPE) \citep{shaw2018selfattentionrelativepositionrepresentations}. \citet{DBLP:conf/iclr/YunBRRK20} defines the function computed by a Transformers with positional encodings as
\[ {\cal T}^{h,m,r}_P = \{g_p(X) =  g(X + E) | g \in {\cal T}^{h,m,r} \ \mbox{and} \ E \in \mathbb{R}^{d \times n} \} \]
There is an analogous theorem for sequence-to-sequence function approximation including the positional encodings, where ${\cal F}_{CD}$ is the set of all continuous functions that map a compact domain in $\mathbb{R}^{n \times d}$. 
\begin{theorem}\citep{DBLP:conf/iclr/YunBRRK20} \label{llmthm}
Let $1 \leq p < \infty$ and $\epsilon > 0$, then for any $f \in {\cal F}_{CD}$, there exists a Transformer network $g \in T^{2,1,4}_P$ such that $d_P(f,g) \leq \epsilon$. 
\end{theorem}
We can restate these results in the categorical framework using the concept of dense functors \citep{richter2020categories} to express the property that Transformer computed functions are dense in the space of all functions in $\mathbb{R}^{d \times n}$ with compact support. We show next how to use categorical abstractions to generalize these results, and construct new Transformer architectures. 

\subsection{Dense Functors} 

We want to categorically capture the property expressed by Theorem~\ref{llmthm} above: the space of all compact functions on $\mathbb{R}^{d \times n}$ is dense in functions representable by Transformers.  We use the concept of dense functors \footnote{For a more detailed introduction, see the web page \url{https://ncatlab.org/nlab/show/dense+functor}.}

\begin{definition}
A functor $i: {\cal S} \rightarrow {\cal C}$ is {\em dense} if every object $c \in {\cal C}$ can be written as the colimit 

\[ \lim_\rightarrow \{ i/c  \xrightarrow[]{Pr} {\cal S}  \rightarrow {\cal C} \} \]
where $\lim_\rightarrow$ is the colimit, $i/C$ is the comma category defined by the functor $i$ and the object $c \in {\cal C}$, $Pr$ is the forgetful functor mapping the comma category $i/C$ onto ${\cal S}$. 
\end{definition}

To build a bit of intuition, note that every set can be constructed as a union of single element sets, which categorically speaking, are just arrows of the form $x: \{ \bullet \} \rightarrow X$, where $\{ \bullet \}$  is the single element set, and $x \in X$. Expressed in the above form, we would say the singleton category is a {\em dense subcategory} of the category of all sets.  We can then state Theorem~\ref{llmthm} in category-theoretic terms as follows: 

\begin{theorem}
    The category of Transformer-representable functions ${\cal T}^{h,m,r}$ is a dense subcategory of the category ${\cal F}_{CD}$ of all functions on $\mathbb{R}^{d \times n}$ with compact support. 
\end{theorem}

{\bf Proof:} The proof is straightforward from the proof of Theorem 2 given in \citep{DBLP:conf/iclr/YunBRRK20}, with the additional definitions required to define the underlying categories. We are simplifying notation a bit here by defining the category of all Transformer-representable functions as ${\cal T}^{h,m,r}$, when what we mean is that each object in this category is such a function, and the arrows of this category are given as commutative diagrams (see Definition~\ref{llmcat-defns}), and similarly for the category category ${\cal F}_{CD}$ of all functions on $\mathbb{R}^{d \times n}$ with compact support. $\qed$

The key idea behind our topos-theoretic construction of architectures for LLMs is that a category whose objects are functions on sets is a topos. The proof of this result is given in standard books \cite{goldblatt:topos}. In the discussion that follows, we will simply view a transformer by its induced function, and implicitly invoke this density theorem to appeal to the case that (co)limits exist precisely because they exist in the category of functions on sets, and that Transformers are dense in this space. This simplification will reduce the length of the proofs considerably, and it is our intention here to communicate the main ideas with as little technical obfuscation as possible. 

\section{Diagrams and Functors in Categories}

We introduce a new way to design LLM architectures as {\em diagrams} of functors. A {\em functor} $F: {\cal C} \rightarrow {\cal D}$ maps the objects $c \in {\cal C}$ to $Fc \in {\cal D}$, as well as each arrow $f: c \rightarrow c' \in {\cal C}$ to $Ff: Fc \rightarrow Fd \in {\cal D}$. Thus, functors are defined by an {\em object function} and an {\em arrows function}.  A {\em diagram} of {\em shape} {\cal J} over category ${\cal C}$ is defined as the functor $F: {\cal J} \rightarrow {\cal C}$. The daisy chaining architecture shown in Figure~\ref{fig:llm-arch} can be abstractly defined as the diagram $DC: {\cal J} \rightarrow {\cal C}^\rightarrow_T$, where the indexing category ${\cal J}$ is just $\bullet \rightarrow \bullet \rightarrow \bullet \ldots$, and ${\cal C}^\rightarrow_T$ is the category of Transformer models.  Examples of diagrams that lead to novel LLM architectures are given below. In the next section, we discuss how to "solve" diagrams.  

\begin{enumerate}
    \item {\bf Pullback diagram}: The pullback diagram is defined as ${\cal J} = \bullet \rightarrow \bullet \leftarrow \bullet$. This architecture defines two LLMs that map to the same co-domain output sequence. 

    \item {\bf Pushforward diagram}: The pushforward diagram is defined as ${\cal J} = \bullet \leftarrow \bullet \rightarrow \bullet$. This architecture defines two LLMs that map from the same domain sequence. 

    \item {\bf Equalizer diagram}: The equalizer diagram is defined as  ${\cal J} = \bullet \rightarrow \bullet \twoarrows[0.4ex]{\rightarrow}{\rightarrow}  \bullet$. This architecture is a generalization of the mixture of LLM model. 

     \item {\bf Co-Equalizer diagram}: The co-equalizer diagram is defined as ${\cal J} = \bullet \twoarrows[0.4ex]{\rightarrow}{\rightarrow}  \bullet \rightarrow \bullet$. This architecture is like a ``dual" of the mixture of LLM model. 
     
\end{enumerate}

The ``cube" diagram in Figure~\ref{fig:llm-arch} is an example of an indexing diagram that can be shown to be assembled from a combination of the above building blocks. In general, it can be shown that any diagram can be built as a combination of such elementary building blocks. Of course, it remains to be seen whether these diagrams are actually ``solvable". For example, from the basic properties of a category, we know that daisy chain diagrams $\bullet \rightarrow \bullet \rightarrow \bullet$ are {\em always} solvable, because by definition, the arrows in a category compose. But, an arbitrary category may not have the right properties for the other diagrams shown above. The good news is that all these diagrams are solvable in the category of sets, ${\cal C}_{\bf Sets}$, as well as in other categories, such as topological spaces, groups etc. We need to show precisely what if any of these diagrams is solvable in our category ${\cal C}^\rightarrow_T$ of Transformer models.  To understand how to solve a diagram, and how diagrams lead to novel LLM architectures, we need to introduce the concept of (co)limits and universal properties. 

\section{Natural Transformations and (co)Limits} 

Once we specify an LLM architecture as a functor diagram $F: {\cal J} \rightarrow {\cal C}^\rightarrow_T$, what does it mean to ``solve" it? We use universal constructions from category theory \citep{riehl2017category}. Briefly, the {\em (co)limit} $\lim F$ is an object $c \in {\cal C}^\rightarrow_T$, i.e. an LLM-represented function, that is the ``closest" possible to the diagram $F$ with respect to the morphisms going into (out of) it, respectively, ``measured" by a universal property defined by a {\em natural transformation}. 

Let us first introduce the concept of natural transformation, which are essentially ``arrows" between functors. Given two functors defining LLM architectures of the same shape, say  $F,G: {\cal J} \rightarrow {\cal C}^\rightarrow_T$, the {\em natural transformation} $\eta: F \Rightarrow G$ between architectures $F$ and $G$ is defined as a collection of arrows $\eta_c: Fc \rightarrow Gc$ for each object $c \in {\cal J}$ such that the following diagram commutes, where $f: c \rightarrow c'$ is an arrow in ${\cal J}$. 
 \begin{center}
 \label{nattransf}
%https://q.uiver.app/#q=WzAsNCxbMCwwLCJVIl0sWzAsMiwiViJdLFsyLDAsIlUnIl0sWzIsMiwiViciXSxbMCwxLCJmIl0sWzAsMiwiaCIsMl0sWzIsMywiZiciLDJdLFsxLDMsImciXV0=
\begin{tikzcd}
	Fc && Gc \\
	\\
	Fc' && {Gc'}
	\arrow["\eta_c"', from=1-1, to=1-3]
	\arrow["Ff", from=1-1, to=3-1]
	\arrow["{Gf}"', from=1-3, to=3-3]
	\arrow["\eta_{c'}", from=3-1, to=3-3]
\end{tikzcd}
 \end{center}
For any object $c \in {\cal C}$ and any diagram of shape ${\cal J}$, the {\em constant functor} $c: {\cal J} \rightarrow {\cal C}$ maps every object $j$ of ${\cal J}$ to $c$ and every morphism $f$ in ${\cal J}$ to the identity morphisms $1_c$. We can define a constant functor embedding as the collection of constant functors $\Delta: C \rightarrow {\cal C}^{\cal J}$ that send each object $c$ in ${\cal C}$ to the constant functor at $c$ and each morphism $f: c \rightarrow c'$ to the constant natural transformation, that is, the natural transformation whose every component is defined to be the morphism $f$. 

\begin{definition}
    A {\bf cone over} a diagram $F: {\cal J} \rightarrow {\cal C}$ with the {\bf summit} or {\bf apex} $c \in {\cal C}$ is a natural transformation $\lambda: c \Rightarrow F$ whose domain is the constant functor at $c$. The components $(\lambda_j: c \rightarrow Fj)_{j \in {\cal J}}$ of the natural transformation can be viewed as its {\bf legs}. Dually, a {\bf cone under} $F$ with {\bf nadir} $c$ is a natural transformation $\lambda: F \Rightarrow c$ whose legs are the components $(\lambda_j: F_j \rightarrow c)_{j \in {\cal J}}$.

    % https://q.uiver.app/#q=WzAsMyxbMiwwLCJjIl0sWzAsMiwiRiBqIl0sWzQsMiwiRmsiXSxbMCwxLCJcXGxhbWJkYV9qIl0sWzAsMiwiXFxsYW1iZGFfayIsMl0sWzEsMiwiRiBmIl1d
\[\begin{tikzcd}
	&& c \\
	\\
	{F j} &&&& Fk
	\arrow["{\lambda_j}", from=1-3, to=3-1]
	\arrow["{\lambda_k}"', from=1-3, to=3-5]
	\arrow["{F f}", from=3-1, to=3-5]
\end{tikzcd}\]
    
\end{definition}
Cones under a diagram are referred to usually as {\em cocones}. Using the concept of cones and cocones, we can now formally define the concept of limits and colimits more precisely. 
\begin{definition}
    For any diagram $F: {\cal J} \rightarrow {\cal C}$, there is a functor $\mbox{Cone}(-, F): {\cal C}^{op} \rightarrow \mbox{{\bf Set}}$, which sends $c \in {\cal C}$ to the set of cones over $F$ with apex $c$. Using the Yoneda Lemma (see Supplementary Materials),  a {\bf limit} of $F$ is defined as an object $\lim F \in {\cal C}$ together with a natural transformation $\lambda: \lim F \rightarrow F$, which can be called the {\bf universal cone} defining the natural isomorphism ${\cal C}(-, \lim F) \simeq \mbox{Cone}(-, F)$.  Dually, for colimits, we can define a functor $\mbox{Cone}(F, -): {\cal C} \rightarrow \mbox{{\bf Set}}$ that maps object $c \in {\cal C}$ to the set of cones under $F$ with nadir $c$. A {\bf colimit} of $F$ is a representation for $\mbox{Cone}(F, -)$. Once again, using the Yoneda Lemma, a colimit is defined by an object $\mbox{Colim} F \in {\cal C}$ together with a natural transformation $\lambda: F \rightarrow \mbox{colim} F$, which defines the {\bf colimit cone} as the natural isomorphism $C(\mbox{colim} F, -) \simeq \mbox{Cone}(F, -)$. 
\end{definition}

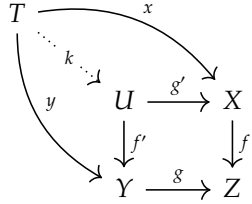
\begin{figure}[h]
\centering
\begin{tikzcd}
  T
  \arrow[drr, bend left, "x"]
  \arrow[ddr, bend right, "y"]
  \arrow[dr, dotted, "k" description] & & \\
    & 
    U\arrow[r, "g'"] \arrow[d, "f'"]
      & X \arrow[d, "f"] \\
& Y \arrow[r, "g"] &Z
\end{tikzcd}
\caption{
 Universal property of pullback mappings. } 
\label{pullback}
\end{figure}

Figure~\ref{pullback}  illustrates the limit of a more complex diagram referred as a {\em {pullback}}, whose diagram is written abstractly as $\bullet \rightarrow \bullet \leftarrow \bullet$.  Note in Figure~\ref{pullback}, the functor maps the diagram $\bullet \rightarrow \bullet \leftarrow \bullet$ to actual objects in the category $Y \xrightarrow[]{g} Z \xleftarrow[]{f} X$. The universal property of the pullback square with the objects $U,X, Y$ and $Z$ implies that the composite mappings $g \circ f'$ must equal $g' \circ f$. In this example, the morphisms $f$ and $g$ represent a {\em {pullback}} pair, as they share a common co-domain $Z$. The pair of morphisms $f', g'$ emanating from $U$ define a {\em {cone}}, because the pullback square ``commutes'' appropriately. Thus, the pullback of the pair of morphisms $f, g$ with the common co-domain $Z$ is the pair of morphisms $f', g'$ with common domain $U$. Furthermore, to satisfy the universal property, given another pair of morphisms $x, y$ with common domain $T$, there must exist another morphism $k: T \rightarrow U$ that ``factorizes'' $x, y$ appropriately, so that the composite morphisms $f' \ k = y$ and $g' \ k = x$. Here, $T$ and $U$ are referred to as {\em cones}, where $U$ is the limit of the set of all cones ``above'' $Z$. If we reverse arrow directions appropriately, we get the corresponding notion of pushforward. 
\subsection{Category of LLMs is (co)Complete}

We now show  the category ${\cal C}^\rightarrow_T$ is complete, that is, it contains all limits and colimits. Hence, all architecture diagrams are ``solvable", meaning that there is an LLM-representable function that defines an object ``closest" to the diagram with respect to the morphisms coming into or out of the diagram.  Our proof hinges on two key properties: in the category ${\cal C}_{\bf Set}$ of sets, all diagrams are solvable, that is limits and colimits exist. Where posssible, we reduce the problem of showing a property to that of the category ${\cal C}_{\bf Set}$. Second, from the density theory shown by \citet{DBLP:conf/iclr/YunBRRK20}, all functions with compact support from $\mathbb{R}^{d \times n}$ to $\mathbb{R}^{d \times n}$  are realizable (to within an arbitrary $\epsilon > 0$) by some Transformer model. 
\begin{theorem}
\label{llmcomplete}
    The  category ${\cal C}^\rightarrow_T$ is {\em (co) complete}, meaning it contains all limits and colimits. 
\end{theorem}
{\bf Proof:}  Formally, this result requires showing that all diagrams, such as pullbacks $\bullet \rightarrow \bullet \leftarrow \bullet$, pushouts $\bullet \leftarrow \bullet \rightarrow \bullet$, (co)equalizer diagrams of the form $\bullet \rightarrow \bullet \twoarrows[0.4ex]{\rightarrow}{\rightarrow}  \bullet$ and $\bullet \twoarrows[0.4ex]{\rightarrow}{\rightarrow}  \bullet \rightarrow \bullet$, respectively, are solvable. For brevity, we will just illustrate the argument for pullback diagrams, and the other arguments are similar. Consider the cube shown in Figure~\ref{fig:llm-arch}. Here, $f$, $g$, and $h$ are the unique functions defining three LLMs, each mapping some input token sequence $I, I', I"$ to some output token sequence $O, O', O"$, respectively. Recall from above that in the new LLM "function objects" category, arrows are commutative diagrams, such as $i,j$ in Figure~\ref{fig:llm-arch}. So, for example, the bottom face of the cube in Figure~\ref{fig:llm-arch} is a commutative diagram, meaning that the relationship $i \circ f = g \circ j$ holds. Similarly,  the arrows $p$, from  I" to  I', and arrow $q$ from O" to O' ensure the right face of the cube is a commutative diagram. The arrow from $P$ to $Q$ exists because looking at the front face of the cube, $Q$ is the pullback of $i$ and $q$, which must exist because we are in the category of sets ${\cal C}_{\bf Set}$, which has all pullbacks. Similarly, the back face of the cube is a pullback of $j$ and $p$, which is again a pullback in ${\cal C}_{\bf Set}$. Summarizing, $\langle u, v \rangle$ and $\langle m, n \rangle$ are the pullbacks of $\langle i, j \rangle$ and $\langle p, q \rangle$. The proof that ${\cal C}^\rightarrow_T$ has all pushouts (limits) is similar. $\qed$

\section{LLM Categories form a Topos}

We now show that LLM categories are not just complete, but they have other properties that make them into a category called a topos \citep{maclane:sheaves,Johnstone:topostheory} that is a ``set-like" category with very special properties, which we will explore in the rest of the paper.  A topos generalizes all common operations on sets. The concept of subset is generalized to a {\em subobject classifier} in a topos.  To help build some intuition, consider how to define subsets without ``looking inside" a set. Essentially, a subset $S$ of some larger set $T$ can be viewed as a ``monic arrow", that is, an injective (or 1-1) function $f: S \hookrightarrow T$. Our approach builds on this abstraction to define a category ${\cal C}^\rightarrow_T$ whose objects are LLMs, and whose arrows map one LLM into another, such as by fine tuning or using RLHF etc. 

\begin{definition}\citep{maclane:sheaves}
    An {\bf elementary topos} is a category ${\cal C}$ that has all (i) limits and colimits, (ii) has exponential objects, and (iii) a subobject classifier.
\end{definition}
\subsection{Subobject Classifiers}
Next, we illustrate the concept of subobject classifiers, and then instantiate this concept for LLMs. A subobject classifier is a generalization of the concept of subset. We can assume that the category has all finite limits, since we already showed that in Theorem~\ref{llmcomplete}. In what follows, a {\em monic} arrow, denoted by $\rightarrowtail$,  means an injective $1-1$ function, such as the mapping from a subset $A \subseteq B$ to the larger set $B$.  We also use ${\bf 1}$ to denote the {\em terminal} object of a category, which has the property that every other object has a unique arrow defined into it. For the category ${\cal C}_{\bf Set}$, the single point set, denoted by $\{ * \}$ is terminal. 
\begin{definition}
    In a category ${\cal C}$ with finite limits, a {\bf subobject classifier} is a ${\cal C}$-object $\Omega$, and a monic ${\cal C}$-arrow ${\bf true}: {\bf 1} \rightarrow \Omega$, such that to every other monic arrow $S \hookrightarrow X$ in ${\cal C}$, there is a unique arrow $\phi$ that forms the following pullback square: 
% https://q.uiver.app/#q=WzAsNCxbMCwwLCJTIl0sWzMsMCwie1xcYmYgMX0iXSxbMCwyLCJYIl0sWzMsMiwie1xcYmYgMn0iXSxbMCwyLCJtIiwwLHsic3R5bGUiOnsidGFpbCI6eyJuYW1lIjoibW9ubyJ9fX1dLFswLDFdLFsxLDMsIntcXGJmIHRydWV9IiwxLHsic3R5bGUiOnsidGFpbCI6eyJuYW1lIjoibW9ubyJ9fX1dLFsyLDMsIlxccGhpX1MiLDEseyJzdHlsZSI6eyJib2R5Ijp7Im5hbWUiOiJkYXNoZWQifX19XV0=
\[\begin{tikzcd}
	S &&& {{\bf 1}} \\
	\\
	X &&& \Omega
	\arrow["m", tail, from=1-1, to=3-1]
	\arrow[from=1-1, to=1-4]
	\arrow["{{\bf true}}"{description}, tail, from=1-4, to=3-4]
	\arrow["{\phi}"{description}, dashed, from=3-1, to=3-4]
\end{tikzcd}\]  
\end{definition}
This commutative diagram enforces a condition that every monic arrow $m$ (i.e., every $1-1$ function) that maps a ``sub"-object $S$ to an object $X$ must be characterizable in terms of a ``pullback", a particular type of universal property that is a special type of a limit. For the category ${\cal C}_{\bf Set}$ of sets, a subobject classifier is the characteristic (Boolean-valued) function $\phi$ that defines subsets. In general, the subobject classifier $\Omega$ is not Boolean-valued, and requires using intuitionistic logic through a Heyting algebra.  This definition can be rephrased as saying that the subobject functor is representable. In other words, a subobject of an object $x$ in a category ${\cal C}$ is an equivalence class of monic arrows $m: S \rightarrowtail  x$. 
 \begin{theorem}
 \label{llmtopos}
     The category ${\cal C}^\rightarrow_T$  forms a topos. 
 \end{theorem}
 {\bf Proof:} The proof essentially involves checking each of the conditions in the above definition of a topos. We will focus on the construction of the subobject classifier and of exponential objects below, since we have already shown above in Theorem~\ref{llmcomplete} that the LLM category has all (co)limits. We prove each of these constructions in the next two sections. $\qed$

 \subsection{Subobject Classifiers for LLMs}

First, we need to define what a ``subobject" is in the category ${\cal C}^\rightarrow_T$. Note that LLMs can abstractly be defined as functions   $f: I \rightarrow O$,  and $g: I' \rightarrow O'$ etc. Here, let us assume that the LLM $M$ that defines $f$ is a {\em submodel} of the LLM $N$ that induces $g$. We can denote that by defining a commutative diagram as shown below. Note here that  $i$ and $j$ are monic arrows. 
\begin{center}
\label{gdcsubobj}
% https://q.uiver.app/#q=WzAsNCxbMCwwLCJVIl0sWzIsMCwiVSciXSxbMCwyLCJWIl0sWzIsMiwiViciXSxbMCwxLCJpIiwwLHsic3R5bGUiOnsidGFpbCI6eyJuYW1lIjoibW9ubyJ9fX1dLFswLDIsImYiLDJdLFsxLDMsImciXSxbMiwzLCJqIiwyLHsic3R5bGUiOnsidGFpbCI6eyJuYW1lIjoibW9ubyJ9fX1dXQ==
\begin{tikzcd}
	I && {I'} \\
	\\
	O && {O'}
	\arrow["i", tail, from=1-1, to=1-3]
	\arrow["f"', from=1-1, to=3-1]
	\arrow["g", from=1-3, to=3-3]
	\arrow["j"', tail, from=3-1, to=3-3]
\end{tikzcd}
\end{center} 
\begin{figure}
    \centering
    \includegraphics[width=0.6\linewidth]{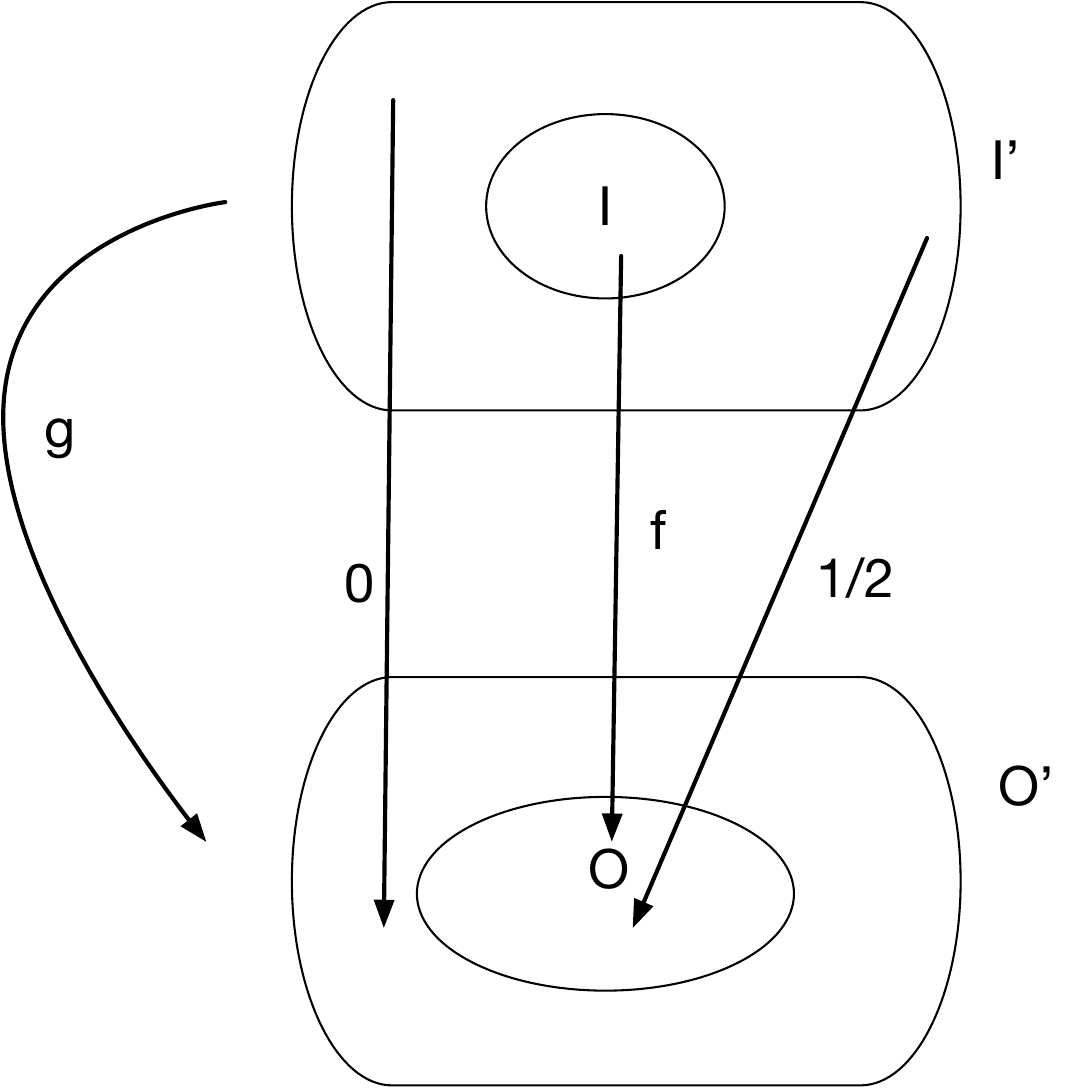}
    \caption{The subobject classifier $\Omega$ for the topos category ${\cal C}^\rightarrow_T$ of LLMs. }
    \label{subobj}
\end{figure}
Let us examine Figure~\ref{subobj} to understand the design of subobject classifiers for the category ${\cal C}^\rightarrow_T$. An element $x \in I'$, which is a particular input sequence, can be classified in three ways by defining a (non-Boolean!) characteristic function $\psi$:
\begin{enumerate}
    \item $x \in I$ -- here we set $\psi(x) = {\bf 1}$. 
    \item $x \notin I$ but $g(x) \in O'$ -- here we set $\psi(x) = {\bf \frac{1}{2}}$.
    \item $x \notin I$ and $g(x) \notin O$ -- we denote this by $\psi(x) = {\bf 0}$. 
\end{enumerate}

\begin{figure}
    \centering
    \includegraphics[width=0.6\linewidth]{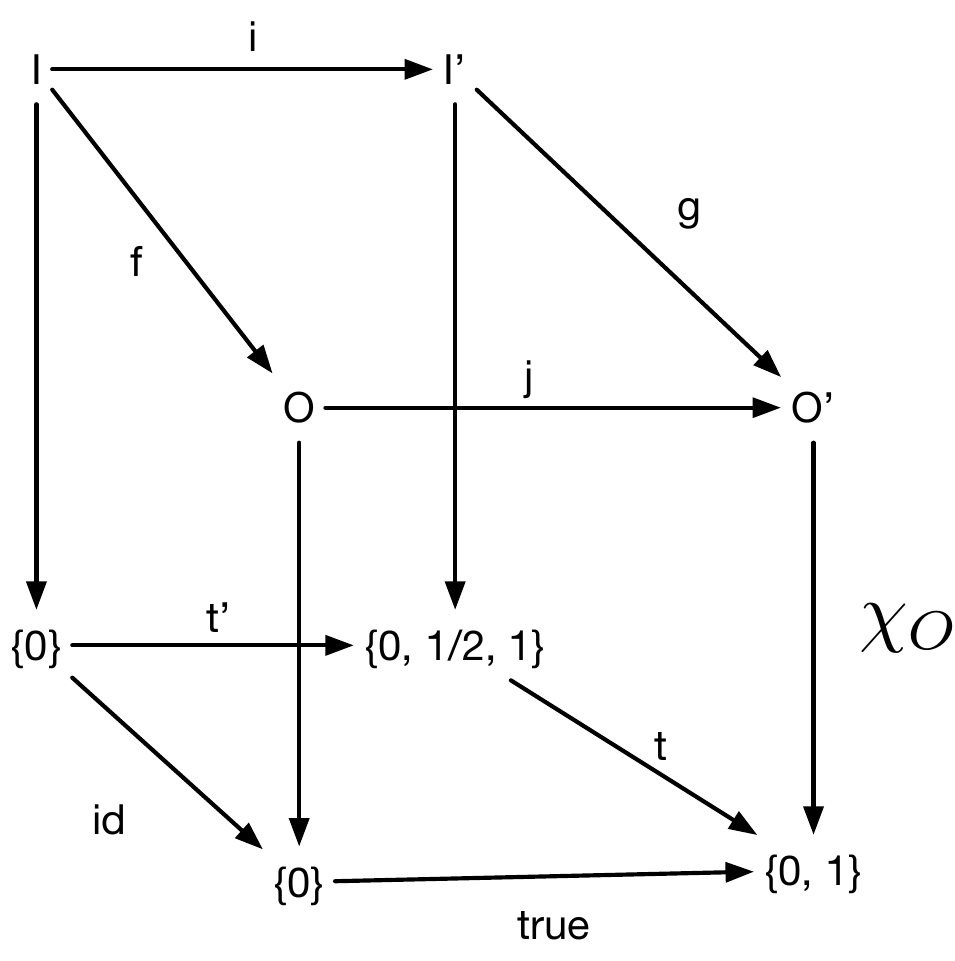}
    \caption{The subobject classifier $\Omega$ for the topos category ${\cal C}^\rightarrow_T$ of LLMs is shown on the bottom face of this cube. }
    \label{subobj-classifier}
\end{figure}

The subobject classifier is illustrated as the bottom face of the cube shown in Figure~\ref{subobj-classifier}: 
\begin{itemize}
    \item ${\bf true}(0) = t'(0) = {\bf 1}$ 
    \item ${\bf t}: {\bf \{0, \frac{1}{2}, 1 \} \rightarrow \{0, 1 \}}$, where ${\bf t(0) = 0}, {\bf t(1) = t(\frac{1}{2}) = 1}$. 
    \item $\chi_O$ is the characteristic function of the output set $O$. 
    \item The base of the cube in Figure~\ref{subobj-classifier} displays the subobject classifier ${\bf T: 1 \rightarrow \Omega}$, where ${\bf T} = \langle {\bf t', true} \rangle$ that maps ${\bf 1 = id_{\{0\}}}$ to $\Omega = {\bf t}: \{0, \frac{1}{2}, 1 \} \rightarrow \{0, 1\}$.
\end{itemize}

This proves that the subobject classifier exists for the Transformer category ${\cal C}^\rightarrow_T$, and it is not Boolean, but has three values of ``truth" , corresponding to the three types of classifications of monic arrows (in regular set theory, subobject classifiers are Boolean: either an element of a parent set is in a subset, or it is not).  

\subsection{Exponential Objects in LLMs}

To complete the proof of Theorem~\ref{llmtopos}, we need to prove also that the category ${\cal C}^\rightarrow_T$ has ``exponential objects". Given two LLM-realized functions $f: I \rightarrow O$, and $g: I' \rightarrow O'$, we define the LLM exponentiated function $g^f: E \rightarrow F$, where $F = O'^O$ is the regular exponential object in the category ${\cal C}_{\bf Set}$ (i.e., all functions from the set $O$ to $O'$),  and $E$ is the collection of all arrows from LLM function $f$ to LLM function $g$ in the category ${\cal C}^\rightarrow_T$,  which can be written more precisely as
\[ E = \{ \langle h, k \rangle | h, k \ \ \mbox{are arrows in the diagram below} \}\]
\begin{center}
%\label{expo-llm}
%https://q.uiver.app/#q=WzAsNCxbMCwwLCJVIl0sWzAsMiwiViJdLFsyLDAsIlUnIl0sWzIsMiwiViciXSxbMCwxLCJmIl0sWzAsMiwiaCIsMl0sWzIsMywiZiciLDJdLFsxLDMsImciXV0=
\begin{tikzcd}
	I && {I'} \\
	\\
	O && {O'}
	\arrow["h"', from=1-1, to=1-3]
	\arrow["f", from=1-1, to=3-1]
	\arrow["{g}"', from=1-3, to=3-3]
	\arrow["k", from=3-1, to=3-3]
\end{tikzcd}
 \end{center}
 and $g^f(\langle h, k \rangle) = k$. First, we define the ``product" object of $g^f$ and $f$ in the LLM category ${\cal C}^\rightarrow_T$ as the product map $g^f \times f: E \times I \rightarrow F \times O$. To show that we have defined a genuine exponential object, we need to demonstrate an ``evaluation" map $g^f \times f \rightarrow g$. The evaluation arrow from $g^f \times f$ to $g$ is the pair $\langle u, v \rangle$ defined by the following commutative diagram:
\begin{center}
 \label{expo-llm}
%https://q.uiver.app/#q=WzAsNCxbMCwwLCJVIl0sWzAsMiwiViJdLFsyLDAsIlUnIl0sWzIsMiwiViciXSxbMCwxLCJmIl0sWzAsMiwiaCIsMl0sWzIsMywiZiciLDJdLFsxLDMsImciXV0=
\begin{tikzcd}
	E \times I && {I'} \\
	\\
	F \times O && {O"}
	\arrow["u"', from=1-1, to=1-3]
	\arrow["g^f \times f", from=1-1, to=3-1]
	\arrow["{g}"', from=1-3, to=3-3]
	\arrow["k", from=3-1, to=3-3]
\end{tikzcd}
 \end{center}
Here, $v$ is the usual evaluation arrow in the category ${\cal C}_{\bf Set}$ of sets, and $u$ maps $\langle \langle h, k \rangle, x \rangle$ to output $h(x)$.

\section{Computational Realization of LLM Topos}  

\label{backprop}

To help ground out the abstract concepts introduced above, we briefly describe how to integrate our topos theory designed LLM architectures with deep learning  \citep{deeplearningreview-2009}. Several categorical deep learning frameworks have been  been published \citep{DBLP:conf/lics/FongST19,gavranović2024positioncategoricaldeeplearning,mahadevan2024gaiacategoricalfoundationsgenerative}. We build on the functorial view of backpropagation introduced in \citep{DBLP:conf/lics/FongST19},  using which we will explore the solution to a more fleshed diagram $F: {\cal J} \rightarrow {\cal S}^\rightarrow_T \rightarrow {\tt Learn}$, where the category ${\tt Learn}$ is a category of compositional learners. 

\subsection{Category {\tt Learn} of Compositional Learners}

We review the work of \citep{DBLP:conf/lics/FongST19}, which models backpropagation as a functor. In particular, they define a category ${\tt Learn}$ of compositional learners as follows. 

\begin{figure}
    \centering
    \includegraphics[scale=0.3]{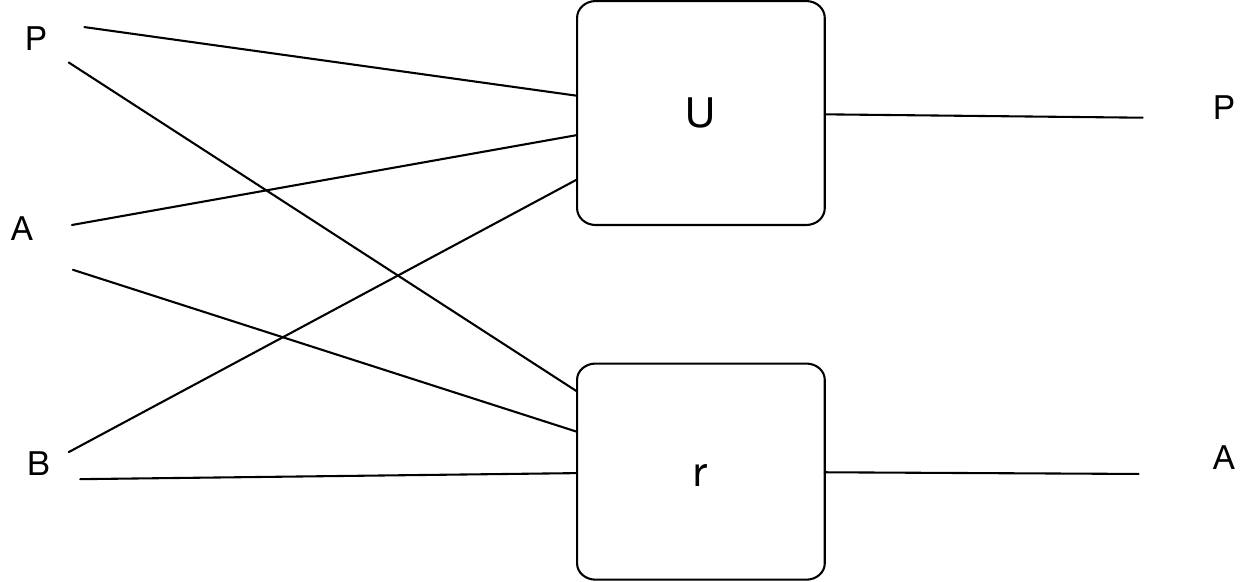}
    \caption{A learner in the symmetric monoidal category {\tt Learn} is defined as a morphism.}
    \label{learncat}
\end{figure}

\begin{definition}\citep{DBLP:conf/lics/FongST19}
    The symmetric monoidal category {\bf Learn} is defined as a collection of objects that define sets, and a collection of an equivalence class of learners. Each learner is defined by the following $4$-tuple (see Figure~\ref{learncat}). 
    
\begin{itemize}
    \item A parameter space $P$

    \item An implementation function $I: P \times A \rightarrow B$

    \item An update function $U: P \times A \times B \rightarrow P$

    \item A request function $r: P \times A \times B \rightarrow A$
\end{itemize}
Two learners $(P, I, U, R)$ and $(P', I', U', r')$ are equivalent if there is a bijection $f: P \rightarrow P'$ such that the following identities hold for each $p \in P, a \in A$ and $b \in B$. 
\begin{itemize}
    \item $I'(f(p), a) = I(p, a)$. 

    \item $U'(f(p), a, b) = f(U(p, a, b))$. 

    \item $r'(f(p), a, b) = r(p, a, b)$
\end{itemize}
    
\end{definition}

\begin{figure}[t]
    \centering
    \includegraphics[scale=0.25]{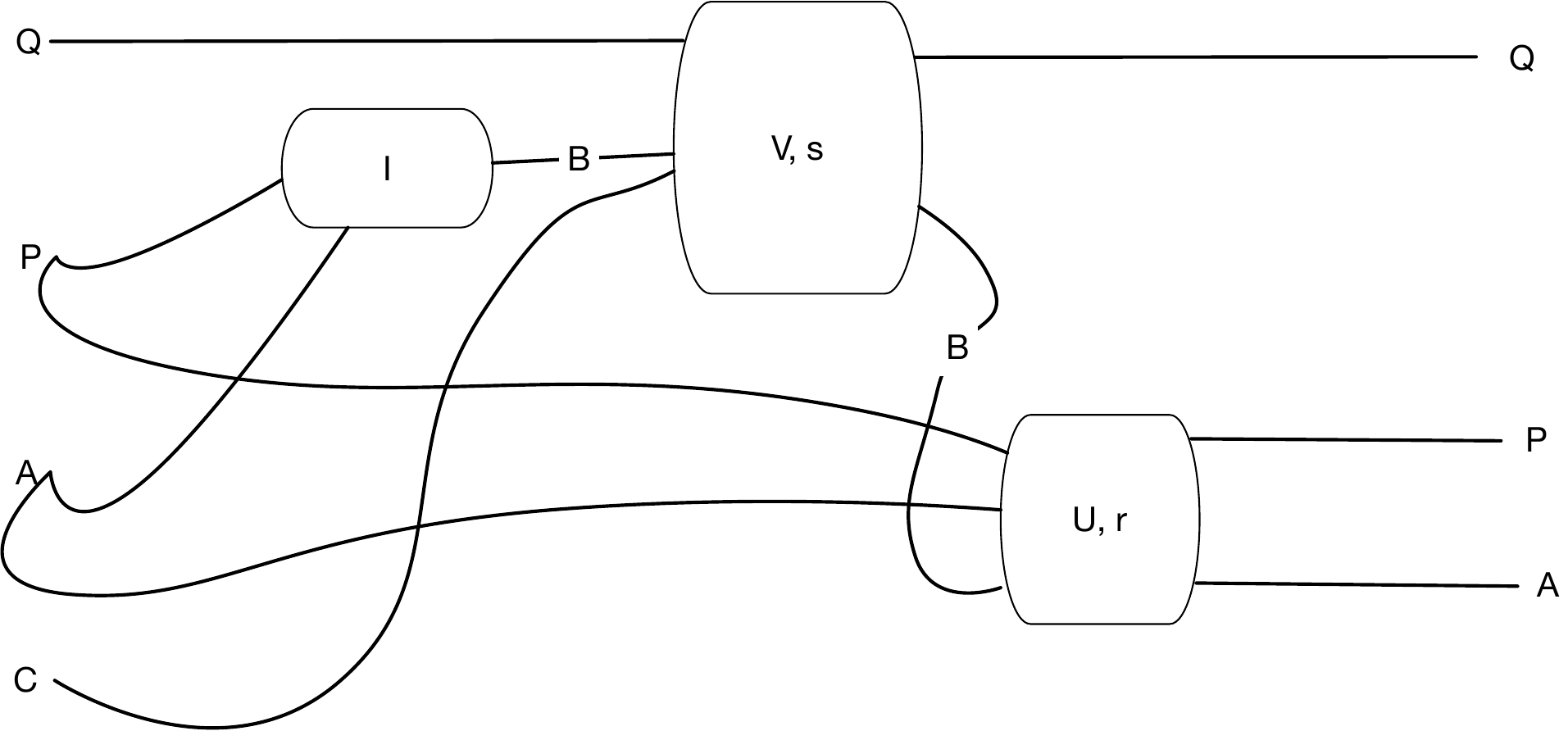}
    \includegraphics[scale=0.25]{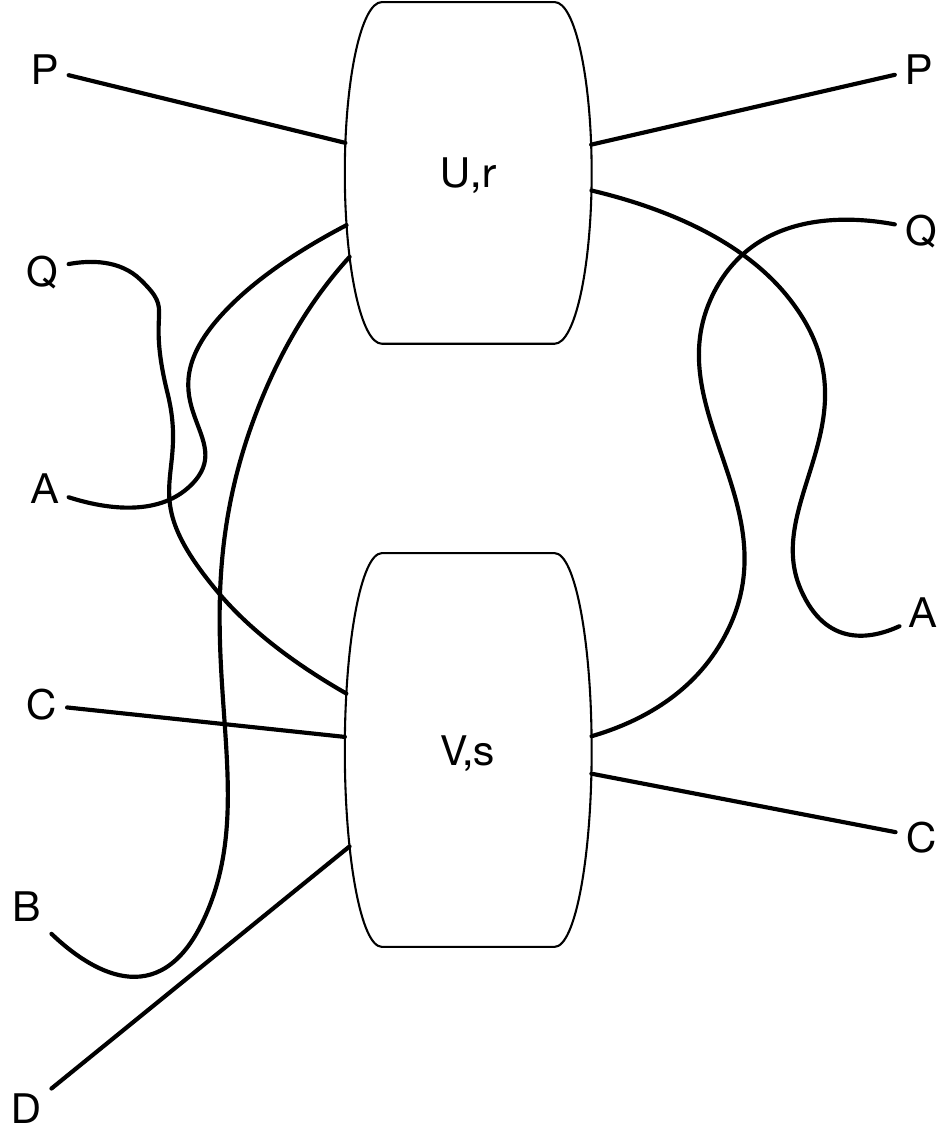}
    \caption{Sequential and parallel composition of two learners in the symmetric monoidal category {\tt Learn}.}
    \label{seqlearn}
\end{figure}

\citep{DBLP:conf/lics/FongST19} show that each learner can be combined in sequentially and in parallel (see Figure~\ref{seqlearn}). We can use these procedures to implement any of our topos-theoretic LLM architectures. For example,  the daisy-chained sequential diagram $\bullet \rightarrow \bullet$, maps into the following structure in ${\tt Learn}$: 
\[ A \xrightarrow[]{(P, I, U, r)} B \xrightarrow[]{(Q, J, V, s)} C \]
The composite learner $A \rightarrow C$ is defined as $(P \times Q, I \cdot J, U \cdot V, r \cdot s)$, where the composite implementation function is
\[ (I \cdot J)(p, q, a) \coloneqq J(q, I(p, a)) \]
and the composite update function is 

\[ U \cdot V(p, q, a, c) \coloneqq \left( U(p, a, s(q, I(p, a), c) \right), V(q, I(p, a), c) \]

and the composite request function is 

\[ (r \cdot s)(p, q, a, c) \coloneqq r(p, a, s(q, I(p, a), c)). \]

\subsection{LLM Topos Implementation as a Functor}

We can now define the LLM topos implementation as a functor $F: {\cal J} \rightarrow {\cal C}^\rightarrow_T \rightarrow {\tt Learn}$, where the first step has already been defined previously in the paper. Note that the category {\tt Learn} is ambivalent as to what particular learning method is used. It could be backpropagation or a zeroth-order stochastic approximation method like {\em random directions} \citep{kushner2003stochastic}. We can define a functor from the LLM category ${\cal C}^\rightarrow_T$  to the category {\tt Learn} that factors through the category {\tt Param}.
% https://q.uiver.app/#q=WzAsMyxbNCwwLCJMZWFybiJdLFswLDAsIk5OZXQiXSxbMiwxLCJQYXJhbSJdLFsxLDIsIkYiXSxbMiwwLCJMX3tcXGVwc2lsb24sIGV9Il0sWzEsMF1d
\[\begin{tikzcd}
	{\cal C}^\rightarrow_T &&&& Learn \\
	&& Param
	\arrow["F_P", from=1-1, to=2-3]
	\arrow["{L_{\epsilon, e}}", from=2-3, to=1-5]
	\arrow[from=1-1, to=1-5]
\end{tikzcd}\]
\begin{definition}
    The category {\tt Param} defines a strict symmetric monoidal category whose objects are $\mathbb{R}^{n \times d}$ token sequences, and whose morphisms are equivalence classes of LLM-defined  functions $f: \mathbb{R}^{n \times d} \rightarrow \mathbb{R}^{n \times d}$. In particular, $(P, I)$ defines a Euclidean space $P$ and $I: P \times A \rightarrow B$ defines a differentiable parameterized function $A \rightarrow B$. Two such pairs $(P, I), (P', I')$ are considered equivalent if there is a differentiable bijection $f: P \rightarrow P'$ such that for all $p \in P$, and $a \in A$, we have that $I'(f'(p),a) = I(p,a)$. The composition of $(P, I): \mathbb{R}^{d \times n} \rightarrow \mathbb{R}^{d \times n}$ and $(Q, J): \mathbb{R}^{d \times n} \rightarrow \mathbb{R}^{d \times n}$ is given as 

    \[ (P \times Q, I \cdot J) \ \ \ \mbox{where} \ \ \ (I \cdot J)(p, q, a) = J(q, I(p, a)) \]

    To model parallel composition of LLMs, we use the monoidal product of objects $\mathbb{R}^{d \times n}$ and $\mathbb{R}^{d \times n}$ giving the object $\mathbb{R}^{2d+2n}$, whereas the monoidal product of morphisms $(P, I): \mathbb{R}^{d \times n} \rightarrow \mathbb{R}^{d \times n}$ and $(Q, J): \mathbb{R}^{d \times n} \rightarrow \mathbb{R}^{d \times n}$ is given as $(P \times Q, I \parallel J)$, where 

    \[ (I \parallel J) (p, q, a, c)  = \left( I(p, a), J(q, c) \right) \]

\end{definition}

The backpropagation algorithm can itself be defined as a functor over symmetric monoidal categories

\[ L_{\epsilon, e}: {\tt Param} \rightarrow {\tt Learn}\]

where $\epsilon > 0$ is a real number defining the learning rate for backpropagation, and $e(x,y): \mathbb{R} \times \mathbb{R} \rightarrow \mathbb{R}$ is a differentiable error function such that $\frac{\partial e}{\partial x}(x_0, -)$ is invertible for each $x_0 \in \mathbb{R}$. This functor essentially defines an update procedure for each parameter in a compositional learner. In other words,  the functor $L_{\epsilon, e}$ defined by backpropagation sends each parameterized function $I: P \times A \rightarrow B$ to the learner $(P, I, U_I,r_I)$

\[ U_I(p, a, b) \coloneqq p - \epsilon \nabla_p E_I(p, a, b) \]

\[ r_I(p, a, b) \coloneqq f_a(\nabla_a E_I(p, a, b)) \]

where $E_I(p, a, b) \coloneqq \sum_j e(I_j(p, a), b_j)$ and $f_a$ is a component-wise application of the inverse to $\frac{\partial e}{\partial x}(a_i, -)$ for each $i$. A detailed empirical evaluation of our topos theory LLM framework will be reported in future work. 

\section{Internal logic of a Large Language Model}

Given our result that the category of LLMs forms a topos, it follows that there is an internal logical language for this category, building on the property that all toposes have such an internal logic. This intriguing prospect makes it possible to reason about LLMs in a novel way, and a full discussion of this idea requires a further paper. But, we give the basic introduction here of what this internal logic is, and what its semantics means. 

\subsection{Local Set Theories} 

\label{lst}

We define formally what an internal local set theory is, and how it can be associated with an externally defined LLM topos category. Our discussion draws from standard textbook treatments, including \citep{maclane:sheaves,Johnstone:topostheory}. We first define local set theories, and then define the Mitchell-B\'enabou internal language of a topos category and specify its Kripke-Joyal semantics.

It is well-understood that properties of sets can be expressed as statements in first-order logic. For example, the following logical statement expresses a property of real numbers: 

\[ \forall x \ \exists y \ \ \ x < y \ \ \ x, y \in \mathbb{R} \]

namely that there does not exist a largest real number. In interpreting such logical statements, every variable $x, y, \ldots $ must be assigned a real number, and has to be interpreted as either ``free" or ``bound" by a quantifier. The above expression has no free variables. Each logical connective, such as $\leq$ must be also given an interpretation. The entire expression has to be assigned a ``truth value" in terms of whether it is true or false. In the development of the internal language associated with a topos, we will see that truth values are not binary, and can take on many possible values. In a presheaf category ${\bf Sets}^{{\cal C}^{op}}$, the subobject classifier $\Omega(C)$ of an object is defined as the partially ordered set of all subobjects, and its ``truth" value is not binary! It is possible to define a ``local set theory" that can be formulated without making any reference at all to sets, but merely as an axiomatic system over a set of abstract types, which will be interpreted in terms of the objects of a topos category below. We briefly sketch out the elements of a local set theory, and refer the reader to the details in \citep{bell}. 

A {\em local set theory} is defined as a language ${\cal L}$ specified by the following classes of symbols: 

\begin{enumerate}
    \item Symbols ${\bf 1} $ and $\Omega$ representing the {\em unity} type and {\em truth-value} type symbols. 

    \item A collection of symbols ${\bf A}, {\bf B}, {\bf C}, \ldots $ called {\em ground type symbols}. 

    \item A collection of symbols ${\bf f}, {\bf g}, {\bf h}, \ldots$ called {\em function} symbols. 
\end{enumerate}

We can use an inductive procedure to recursively construct {\bf type symbols} of ${\cal L}$ as follows: 

\begin{enumerate}
    \item  Symbols ${\bf 1} $ and $\Omega$ are type symbols.

    \item Any ground type symbol is a type symbol. 

    \item If ${\bf A}_1, \ldots, {\bf A}_n$ are type symbols, so is their product ${\bf A}_1 \times \ldots {\bf A}_n$, where for $n=0$, the type of $\prod_{i=1}^n {\bf A}_i$ is ${\bf 1}$. The product ${\bf A}_1 \times \ldots {\bf A}_n$ has the {\em product type} symbol. 

     \item If ${\bf A}$ is a type symbol, so is ${\bf PA}$. The type ${\bf PA}$ is called the {\em power} type. \footnote{Note that in a topos, these will be interpreted as {\em power objects}, generalizing the concept of power sets.}
\end{enumerate}

For each type symbol ${\bf A}$, the language ${\cal L}$ contains a set of {\em variables} $x_{\bf A}, y_{\bf A}, z_{\bf A}, \ldots$. In addition, ${\cal L}$ contains the distinguished ${\bf *}$ symbol. Each function symbol in ${\cal L}$  is assigned a {\em signature} of the form ${\bf A} \rightarrow {\bf B}$. \footnote{In a topos, these will correspond to arrows of the category.} We can define the {\em terms} of the local set theory language ${\cal L}$ recursively as follows: 

\begin{itemize}
    \item ${\bf *}$ is a term of type ${\bf 1}$. 

    \item for each type symbol ${\bf A}$, variables $x_{\bf A}, y_{\bf A}, \ldots$ are terms of type ${\bf A}$. 

    \item if ${\bf f}$ is a function symbol with signature ${\bf A} \rightarrow {\bf B}$, and $\tau$ is a term of type ${\bf A}$, then ${\bf f}(\tau)$ is a term of type ${\bf B}$. 

    \item If $\tau_1, \ldots, \tau_n$ are terms of types ${\bf A}_1, \ldots, {\bf A}_n$, then $\langle \tau_1, \ldots \tau_n \rangle$ is a term of type ${\bf A}_1 \times \ldots {\bf A}_n$, where if $n=0$, then $\langle \tau_1, \ldots \tau_n \rangle$ is of type ${\bf *}$. 

    \item If $\tau$ is a term of type ${\bf A}_1 \times {\bf A}_n$, then for $1 \leq i \leq n$, $(\tau)_i$ is a term of type ${\bf A}_i$. 

    \item if $\alpha$ is a term of type $\Omega$, and $x_{\bf A}$ is a variable of type ${\bf A}$, then $\{x_{\bf A} : \alpha \}$ is a term of type ${\bf PA}$.  

    \item if $\sigma, \tau$ are terms of the same type, $\sigma = \tau$ is a term of type $\Omega$. 

    \item if $\sigma, \tau$ are terms of the types ${\bf A}, {\bf PA}$, respectively, then $\sigma \in \tau$ is a term of type ${\bf \Omega}$. 
\end{itemize}

A term of type ${\bf \Omega}$ is called a {\em formula}. The language ${\cal L}$ does not yet have defined any logical operations, because in a typed language, logical operations can be defined in terms of the types, as illustrated below. 

\begin{itemize}
    \item $\alpha \Leftrightarrow \beta$ is interpreted as $\alpha = \beta$. 

    \item {\bf true} is interpreted as ${\bf *} = {\bf *}$. 

    \item $\alpha \wedge \beta$ is interpreted as $\langle \alpha, \beta \rangle = \langle {\bf true}, {\bf false} \rangle$. 

    \item $\alpha \Rightarrow \beta$ is interpreted as $(\alpha \wedge \beta) \Leftrightarrow \alpha$

    \item $\forall x \ \alpha$ is interpreted as $\{x : \alpha\} = \{x : {\bf true} \}$

    \item ${\bf false}$ is interpreted as $\forall \omega \ \omega$. 

    \item $\neg \alpha$ is interpreted as $\alpha \Rightarrow {\bf false}$.

    \item $\alpha \vee \beta$ is interpreted as $\forall \omega \ [(\alpha \Rightarrow \omega \wedge \beta \Rightarrow \omega) \Rightarrow \omega]$

    \item $\exists x \ \alpha$ is interpreted as $\forall \omega [ \forall x (\alpha \Rightarrow \omega) \Rightarrow \omega ]$

\end{itemize}

Finally, we have to specify the inference rules, which are given in the form of {\em sequents}. We will just sketch out a few, and the rest can be seen in \citep{bell}. A sequent is a formula 

\[ \Gamma: \alpha\]

where $\alpha$ is a formula, and $\Gamma$ is a possibly empty finite set of formulae. The basic axioms include $\alpha: \alpha$ (tautology), $:x_1 = {\bf *}$ (unity), a rule for forming projections of products, a rule for equality, and another for comprehension. Finally, the inference rules are given in the form: 

\begin{itemize}
    \item {\em Thinning:}
    \[
  \begin{prooftree}
    \hypo{\Gamma : \alpha}
    \infer1{\beta, \Gamma: \alpha}
  \end{prooftree}
\]
\item {\em Cut}: 

\[
  \begin{prooftree}
    \hypo{\Gamma : \alpha, \  \ \alpha, \Gamma: \beta}
    \infer1{\Gamma: \beta}
  \end{prooftree}
\]

\item {\em Equivalence}: 

\[
  \begin{prooftree}
    \hypo{\alpha, \Gamma : \beta \ \ \beta, \Gamma: \alpha}
    \infer1{\Gamma: \alpha \Leftrightarrow \beta}
  \end{prooftree} 
\]

\end{itemize}

A full list of inference rules with examples of proofs is given in \citep{bell}. Now that we have the elements of a local set theory defined as shown above, we need to connect its definitions with that of a topos. That is the topic of the next section.

\subsection{Mitchell-B\'enabou Language of a Topos}
\label{mbl}

We now define the Mitchell-B\'enabou language (MBL) associated with any topos category \citep{maclane:sheaves}.  As with the abstract local set theory defined in the previous section, we have to define the types (which will be the objects of a topos), the functions and terms, and give definition of universal and existential quantifiers. We postpone the discussion of the interpretation of this language to the next section. 

Given a topos category ${\cal C}$, we define the types of MBL as the objects of ${\cal C}$. Note that for an LLM category ${\cal C}^\rightarrow_T$, the types will correspond to the LLM-induced functions $f: \mathbb{R}^{d \times n} \rightarrow \mathbb{R}^{d \times n}$. 

For each type $C$ (defined as an object of the topos category ${\cal C}$), like for a local set theory, we assume the existence of variables $x_C, y_C, \ldots$, where each such variable has as its interpretation the identity arrow ${\bf 1}: C \rightarrow C$. Just like for local set theories, we can construct product objects, such as $A \times B \times C$, where terms like $\sigma$ that define arrows are given the interpretation 

\[ \sigma: A \times B \times C \rightarrow D \]

We can inductively define the terms and their interpretations in a topos category as follows (see \citep{maclane:sheaves} for additional details): 

\begin{itemize}
    \item Each variable $x_C$ of type $C$ is a term of type $C$, and its interpretation is the identity $x_C = {\bf 1}: C \rightarrow C$. 

\item Terms $\sigma$ and $\tau$ of types $C$ and $D$ that are interpreted as $\sigma: A \rightarrow C$ and $\tau: B \rightarrow D$ can be combined to yield a term $\langle \sigma, \tau \rangle$ of type $C \times D$, whose joint interpretation is given as 

\[ \langle \sigma p, \tau q \rangle: X \rightarrow C \times D\]

where $X$ has the required projections $p: X \rightarrow A$  and $q: X \rightarrow B$. 

\item Terms $\sigma: A \rightarrow B$ and $\tau: C \rightarrow B$ of the same type $B$ yield a term $\sigma = \tau$ of type $\Omega$, interpreted as 

\[ (\sigma = \tau): W \xrightarrow[]{\langle \sigma p, \tau q \rangle} B \times B \xrightarrow[]{\delta_B} \Omega \]

where $\delta_B$ is the characteristic map of the diagonal functor $\Delta B \rightarrow B \times B$. In the AGI modality for causal inference,  these diagonal maps will correspond to the ``copy" procedure in a topos category of presheaves over Markov categories \citep{Fritz_2020}. 

\item Arrows $f: A \rightarrow B$ and a term $\sigma: C \rightarrow A$ of type $A$ can be combined to yield a term $f \circ \sigma$ of type $B$, whose interpretation is naturally a composite arrow: 

\[ f \circ \sigma: C \xrightarrow[]{\sigma} A \xrightarrow[]{f} B\]

\item For exponential objects, terms $\theta: A \rightarrow B^C$ and $\sigma: D \rightarrow C$ of types $B^C$ and $C$, respectively, combine to give an ``evaluation" map of type $B$, defined as 

\[ \theta (\sigma): W \rightarrow B^C \times C \xrightarrow[]{e} B \]

where $e$ is the evaluation map, and $W$ defines a map $\langle \theta p, \sigma q \rangle$, where once again $p: W \rightarrow A$ and $q: W \rightarrow D$ are projection maps. 

\item Terms $\sigma: A \rightarrow B$ and $\tau: D \rightarrow \Omega^B$ combine to yield a term $\sigma \in \tau$ of type $\Omega$, with the following interpretation: 

\[ \sigma \in \tau: W \xrightarrow[]{\langle \sigma p, \tau q \rangle} B \times \Omega^B \xrightarrow[]{e} \Omega \]

\item Finally, we can define local functions as $\lambda$ objects, such as 

\[ \lambda x_C \sigma: A \rightarrow B^C \]

where $x_C$ is a variable of type $C$ and $\sigma: C \times A \rightarrow B$. 
\end{itemize}

Once again, we can combine terms $\alpha, \beta$ etc. of type $\Omega$ using logical connectives $\wedge, \vee, \Rightarrow, \neg$, as well as quantifiers, to get composite terms, where each of the logical connectives is now defined over the subobject classifier $\Omega$, giving us

\begin{itemize}
    \item $\wedge: \Omega \times \Omega \rightarrow \Omega$ is interpreted as the {\em meet} operation in the partially ordered set of subobjects (given by the Heyting algebra). 

    \item $\vee: \Omega \times \Omega \rightarrow \Omega$ is interpreted as the {\em join} operation in the partially ordered set of subobjects (given by the Heyting algebra). 

    \item $\Rightarrow: \Omega \times \Omega \rightarrow \Omega$ is interpreted as an adjoint functor, as defined previously for a Heyting algebra. 
    
\end{itemize}

We can combine these logical connectives with the term interpretation as arrows as defined earlier in a fairly straightforward way, as described in \citep{maclane:sheaves}. We now turn to the Kripe-Joyal semantics of this language. 

\subsection{Kripke-Joyal Semantics}
\label{kj}

 Let ${\cal C}$ be a topos, and let it possess a Mitchell-B\'enabou language as defined above. How do we define a suitable model for this language? In this section, we define the Kripke-Joyal semantics that provides an interpretation of the Mitchell-B\'enabou language described in the previous section. A more detailed overview  of this topic is given in \citep{maclane:sheaves}.

 For the category ${\cal C}$, and for any object $X$ in ${\cal C}$, define a {\em generalized element} as simply a morphism $\alpha: U \rightarrow X$. We want to specify the semantics of how $U$ supports any formula $\phi(\alpha)$, denoted by $U \Vdash \phi(\alpha)$. We declare that this ``forcing" relationship holds if and only if $\alpha$ factors through $\{x | \phi(x) \}$, where $x$ is a variable of type $X$ (recall that objects $X$ of a topos form its types), as shown in the following commutative diagram. 

 \begin{center}
     % https://q.uiver.app/#q=WzAsNSxbMCwyLCJVIl0sWzIsMCwiXFx7eCB8IFxccGhpKHgpIFxcfSJdLFsyLDIsIlgiXSxbNCwwLCJ7XFxiZiAxfSJdLFs0LDIsIntcXGJmIFxcT21lZ2F9ICJdLFswLDEsIiIsMCx7InN0eWxlIjp7ImJvZHkiOnsibmFtZSI6ImRhc2hlZCJ9fX1dLFswLDIsIlxcYWxwaGEiLDJdLFsxLDIsIiIsMCx7InN0eWxlIjp7InRhaWwiOnsibmFtZSI6Im1vbm8ifX19XSxbMSwzXSxbMyw0LCJ7XFxiZiBUcnVlfSJdLFsyLDQsIlxccGhpKHgpIl1d
\begin{tikzcd}
	&& {\{x | \phi(x) \}} && {{\bf 1}} \\
	\\
	U && X && {{\bf \Omega} }
	\arrow[from=1-3, to=1-5]
	\arrow[tail, from=1-3, to=3-3]
	\arrow["{{\bf True}}", from=1-5, to=3-5]
	\arrow[dashed, from=3-1, to=1-3]
	\arrow["\alpha"', from=3-1, to=3-3]
	\arrow["{\phi(x)}", from=3-3, to=3-5]
\end{tikzcd}
 \end{center}

 Building on this definition, if $\alpha, \beta: U \rightarrow X$ are parallel arrows, we can give semantics to the formula $\alpha = \beta$ by the following statement: 

 \[ U \xrightarrow[]{\langle \alpha, \beta \rangle} X \times X \xrightarrow[]{\delta_X} \Omega\]

following the definitions in the previous section for the composite $\langle \alpha, \beta \rangle$ and $\delta_X$ in MBL. 

We can extend the previous commutative diagram to show that $U \Vdash \alpha = \beta$ holds if and only if $\langle \alpha, \beta \rangle$ factors through the diagonal map $\Delta$: 

\begin{center}
    % https://q.uiver.app/#q=WzAsNSxbMCwyLCJVIl0sWzIsMCwiWCJdLFsyLDIsIlggXFx0aW1lcyBYIl0sWzQsMCwie1xcYmYgMX0iXSxbNCwyLCJ7XFxiZiBcXE9tZWdhfSAiXSxbMCwxLCIiLDAseyJzdHlsZSI6eyJib2R5Ijp7Im5hbWUiOiJkYXNoZWQifX19XSxbMCwyLCJcXGxhbmdsZSBcXGFscGhhLCBcXGJldGEgXFxyYW5nbGUiLDJdLFsxLDIsIlxcRGVsdGEiLDAseyJzdHlsZSI6eyJ0YWlsIjp7Im5hbWUiOiJtb25vIn19fV0sWzEsM10sWzMsNCwie1xcYmYgVHJ1ZX0iXSxbMiw0LCJcXGRlbHRhX3giXV0=
\begin{tikzcd}
	&& X && {{\bf 1}} \\
	\\
	U && {X \times X} && {{\bf \Omega} }
	\arrow[from=1-3, to=1-5]
	\arrow["\Delta", tail, from=1-3, to=3-3]
	\arrow["{{\bf True}}", from=1-5, to=3-5]
	\arrow[dashed, from=3-1, to=1-3]
	\arrow["{\langle \alpha, \beta \rangle}"', from=3-1, to=3-3]
	\arrow["{\delta_x}", from=3-3, to=3-5]
\end
{tikzcd}
\end{center}

Many additional properties can be derived (see \citep{maclane:sheaves}), including the following useful ones. 

\begin{itemize}
    \item {\bf Monotonicity:} If $U \Vdash \phi(x)$, then we can pullback the interpretation through any arrow $f: U' \rightarrow U$ in a topos ${\cal C}$ to obtain $U' \Vdash \phi(\alpha \circ f)$. 
    \begin{center}
        % https://q.uiver.app/#q=WzAsNixbMiwyLCJVIl0sWzQsMCwiXFx7eCB8IFxccGhpKHgpIFxcfSJdLFs0LDIsIlgiXSxbNiwwLCJ7XFxiZiAxfSJdLFs2LDIsIntcXGJmIFxcT21lZ2F9ICJdLFswLDIsIlUnIl0sWzAsMSwiIiwwLHsic3R5bGUiOnsiYm9keSI6eyJuYW1lIjoiZGFzaGVkIn19fV0sWzAsMiwiXFxhbHBoYSIsMl0sWzEsMiwiIiwwLHsic3R5bGUiOnsidGFpbCI6eyJuYW1lIjoibW9ubyJ9fX1dLFsxLDNdLFszLDQsIntcXGJmIFRydWV9Il0sWzIsNCwiXFxwaGkoeCkiXSxbNSwwLCJmIl0sWzUsMSwiIiwwLHsic3R5bGUiOnsiYm9keSI6eyJuYW1lIjoiZGFzaGVkIn19fV1d
\begin{tikzcd}
	&&&& {\{x | \phi(x) \}} && {{\bf 1}} \\
	\\
	{U'} && U && X && {{\bf \Omega} }
	\arrow[from=1-5, to=1-7]
	\arrow[tail, from=1-5, to=3-5]
	\arrow["{{\bf True}}", from=1-7, to=3-7]
	\arrow[dashed, from=3-1, to=1-5]
	\arrow["f", from=3-1, to=3-3]
	\arrow[dashed, from=3-3, to=1-5]
	\arrow["\alpha"', from=3-3, to=3-5]
	\arrow["{\phi(x)}", from=3-5, to=3-7]
\end{tikzcd}
    \end{center}

    \item {\bf Local character:} Analogously, if $f: U' \rightarrow U$ is an epic arrow, then from $U' \Vdash \phi(\alpha \circ f)$, we can conclude $U \Vdash \phi(x)$. 
\end{itemize}

We can summarize the main results of Kripke-Joyal semantics using the following theorem. These give precise semantics for the standard logical connectives, as well as universal and existential quantification in terms of the arrows of a topos category ${\cal C}$. We can specialize these broad results to specific AGI categories in the subsequent sections. 

\begin{theorem}\citep{maclane:sheaves}
    If $\alpha: U \rightarrow X$ is a generalized element of $X$, and $\phi(x)$ and $\psi(x)$ are formulas with a free variable $x$ of type $X$, we can conclude that
    \begin{enumerate}
        \item $U \Vdash \phi(\alpha) \wedge \psi(\alpha)$ holds if $U \Vdash \phi(\alpha)$ and $U \Vdash \psi(\alpha)$. 
        \item $U \Vdash \phi(x) \vee \psi(x)$ holds if there are morphisms $p: V \rightarrow U$ and $q: W \rightarrow U$ such that $p + q: V + W \rightarrow U$ is an epic arrow, and $V \Vdash \phi(\alpha p)$ and $W \Vdash \phi(\alpha q)$. 
        \item $U \Vdash \phi(\alpha) \Rightarrow \psi(\alpha)$ if it holds that for any morphism $p: V \rightarrow U$, where $V \Vdash \phi(\alpha p)$, the assertion $V \Vdash \phi(\alpha p)$  also holds. 

        \item $U \Vdash \neg \phi(\alpha)$ holds if whenever the morphism $p: U \rightarrow V$ satisfies the property $V \Vdash \phi(\alpha p)$, then $V \cong {\bf 0}$. 

        \item $U \Vdash \exists \phi(x,y)$ holds if there exists an epic arrow $p: V \rightarrow U$ and generalized elements $\beta: V \rightarrow Y$ such that $V \Vdash \phi(\alpha p, \beta)$. 

        \item $U \Vdash \forall y \phi(x, y)$ holds if for every object $V$, and every arrow $p: V \rightarrow U$, and every generalized element $\beta: V \rightarrow Y$, it holds that $V \Vdash \phi(\alpha p, \beta)$. 
    \end{enumerate}
\end{theorem}

To understand the significance of this theorem, note that we can now use it to provide rigorous semantics for the LLM topos category ${\cal C}^\rightarrow_T$. 

Summarizing this  section, we began by defining a local set theory of types, within which we were able to state the language ${\cal L}$ and its inference rules. These abstractly characterize what a ``set-like" category should behave as. Subsequently, we showed that the Mitchell-B\'enabou language for a topos is precisely of the form of a local set theory, formalizing the precise way in which a topos is like a category of sets. Finally, we specified the Kripke-Joyal semantics for the Mitchell-B\'enabou internal language of a topos.  

\section{Summary and Future Work}

In this paper, we proposed using topos theory to design novel generative AI architectures (GAIAs), focusing on LLMs as the paradigmatic example. Building on the correspondence between the space of all functions on Euclidean-embedded token sequences with compact support and LLM-representable functions, we showed the category of LLM objects is (co)complete, and also forms a topos. We built on previous theoretical results on the Transformer model, which show that it is a universal sequence-to-sequence function approximator, and dense in the space of all continuous functions on $\mathbb{R}^{d \times n}$ with compact support. Previous studies of large language models (LLMs) have focused on daisy-chained linear architectures or mixture-of-experts. In this paper, we use the theory of categories and functors  to construct much richer LLM architectures based on new types of compositional structures. In particular, these new compositional structures are derived from universal properties of LLM categories, and include {\em pullback}, {\em pushout}, {\em (co) equalizers}, exponential compositions, and subobject classifiers.  We theoretically validate these new compositional structures by showing that the category of LLMs is (co)complete, meaning that all diagrams have solutions in the form of (co)limits.  Building on this completeness result, we then show that the category of LLMs forms a topos, a ``set-like" category, which requires showing the existence of exponential objects as well as subobject classifiers. We use a functorial characterization of 

Several avenues for future work need to be explored, which are briefly discussed below. 

\begin{itemize}
    \item {\bf Implementation}: An obvious question is whether the topos-theoretic LLM architectures actually give superior performance compared to existing daisy-chained and mixture of LLM architectures. This question is clearly a topic for a future experimental paper. 

    \item {\bf Theory:} Existing theoretical results for LLMs are based on the simple daisy-chained architecture. It is an intriguing question whether the enhanced (co)limit and subobject classifer based LLMs will provide any additional theoretical power. This question is also clearly a topic for a future paper. 

    \item {\bf Other generative AI models:} The framework has been described largely for LLMs, but it clearly extends to other generative AI models, such as structured state space models and other types of diffusion models.  This topic seems worth exploring as well in a future paper. 
\end{itemize}

%\bibliography{cc,AuthorKit26/AnonymousSubmission/LaTeX/agicitations}

\newpage 

\begin{thebibliography}{30}
\providecommand{\natexlab}[1]{#1}
\providecommand{\url}[1]{\texttt{#1}}
\expandafter\ifx\csname urlstyle\endcsname\relax
  \providecommand{\doi}[1]{doi: #1}\else
  \providecommand{\doi}{doi: \begingroup \urlstyle{rm}\Url}\fi

\bibitem[Bell(1988)]{bell}
J.~L. Bell.
\newblock \emph{Toposes and Local Set Theories}.
\newblock Dover, 1988.

\bibitem[Bengio(2009)]{deeplearningreview-2009}
Y.~Bengio.
\newblock Learning deep architectures for {AI}.
\newblock \emph{Foundations and Trends in Machine Learning}, 2\penalty0 (1):\penalty0 1--127, 2009.

\bibitem[Bommasani et~al.(2022)Bommasani, Hudson, Adeli, Altman, Arora, von Arx, Bernstein, Bohg, Bosselut, Brunskill, Brynjolfsson, Buch, Card, Castellon, Chatterji, Chen, Creel, Davis, Demszky, Donahue, Doumbouya, Durmus, Ermon, Etchemendy, Ethayarajh, Fei-Fei, Finn, Gale, Gillespie, Goel, Goodman, Grossman, Guha, Hashimoto, Henderson, Hewitt, Ho, Hong, Hsu, Huang, Icard, Jain, Jurafsky, Kalluri, Karamcheti, Keeling, Khani, Khattab, Koh, Krass, Krishna, Kuditipudi, Kumar, Ladhak, Lee, Lee, Leskovec, Levent, Li, Li, Ma, Malik, Manning, Mirchandani, Mitchell, Munyikwa, Nair, Narayan, Narayanan, Newman, Nie, Niebles, Nilforoshan, Nyarko, Ogut, Orr, Papadimitriou, Park, Piech, Portelance, Potts, Raghunathan, Reich, Ren, Rong, Roohani, Ruiz, Ryan, Ré, Sadigh, Sagawa, Santhanam, Shih, Srinivasan, Tamkin, Taori, Thomas, Tramèr, Wang, Wang, Wu, Wu, Wu, Xie, Yasunaga, You, Zaharia, Zhang, Zhang, Zhang, Zhang, Zheng, Zhou, and Liang]{fm}
Rishi Bommasani, Drew~A. Hudson, Ehsan Adeli, Russ Altman, Simran Arora, Sydney von Arx, Michael~S. Bernstein, Jeannette Bohg, Antoine Bosselut, Emma Brunskill, Erik Brynjolfsson, Shyamal Buch, Dallas Card, Rodrigo Castellon, Niladri Chatterji, Annie Chen, Kathleen Creel, Jared~Quincy Davis, Dora Demszky, Chris Donahue, Moussa Doumbouya, Esin Durmus, Stefano Ermon, John Etchemendy, Kawin Ethayarajh, Li~Fei-Fei, Chelsea Finn, Trevor Gale, Lauren Gillespie, Karan Goel, Noah Goodman, Shelby Grossman, Neel Guha, Tatsunori Hashimoto, Peter Henderson, John Hewitt, Daniel~E. Ho, Jenny Hong, Kyle Hsu, Jing Huang, Thomas Icard, Saahil Jain, Dan Jurafsky, Pratyusha Kalluri, Siddharth Karamcheti, Geoff Keeling, Fereshte Khani, Omar Khattab, Pang~Wei Koh, Mark Krass, Ranjay Krishna, Rohith Kuditipudi, Ananya Kumar, Faisal Ladhak, Mina Lee, Tony Lee, Jure Leskovec, Isabelle Levent, Xiang~Lisa Li, Xuechen Li, Tengyu Ma, Ali Malik, Christopher~D. Manning, Suvir Mirchandani, Eric Mitchell, Zanele Munyikwa, Suraj Nair,
  Avanika Narayan, Deepak Narayanan, Ben Newman, Allen Nie, Juan~Carlos Niebles, Hamed Nilforoshan, Julian Nyarko, Giray Ogut, Laurel Orr, Isabel Papadimitriou, Joon~Sung Park, Chris Piech, Eva Portelance, Christopher Potts, Aditi Raghunathan, Rob Reich, Hongyu Ren, Frieda Rong, Yusuf Roohani, Camilo Ruiz, Jack Ryan, Christopher Ré, Dorsa Sadigh, Shiori Sagawa, Keshav Santhanam, Andy Shih, Krishnan Srinivasan, Alex Tamkin, Rohan Taori, Armin~W. Thomas, Florian Tramèr, Rose~E. Wang, William Wang, Bohan Wu, Jiajun Wu, Yuhuai Wu, Sang~Michael Xie, Michihiro Yasunaga, Jiaxuan You, Matei Zaharia, Michael Zhang, Tianyi Zhang, Xikun Zhang, Yuhui Zhang, Lucia Zheng, Kaitlyn Zhou, and Percy Liang.
\newblock On the opportunities and risks of foundation models, 2022.

\bibitem[Bradley et~al.(2022)Bradley, Terilla, and Vlassopoulos]{bradley:enriched-yoneda-llms}
TD. Bradley, J.~Terilla, and Y.~Vlassopoulos.
\newblock An enriched category theory of language: From syntax to semantics.
\newblock \emph{La Matematica}, 1:\penalty0 551--580, 2022.

\bibitem[Chaudhari et~al.(2021)Chaudhari, Mithal, Polatkan, and Ramanath]{chaudhari2021attentivesurveyattentionmodels}
Sneha Chaudhari, Varun Mithal, Gungor Polatkan, and Rohan Ramanath.
\newblock An attentive survey of attention models, 2021.
\newblock URL \url{https://arxiv.org/abs/1904.02874}.

\bibitem[Chiang et~al.(2023)Chiang, Cholak, and Pillay]{transformer-bounds}
David Chiang, Peter Cholak, and Anand Pillay.
\newblock Tighter bounds on the expressivity of transformer encoders.
\newblock In \emph{Proceedings of the 40th International Conference on Machine Learning}, ICML'23. JMLR.org, 2023.

\bibitem[DeepSeek-AI et~al.(2025)DeepSeek-AI, Guo, Yang, Zhang, Song, Zhang, Xu, Zhu, Ma, Wang, Bi, Zhang, Yu, Wu, Wu, Gou, Shao, Li, Gao, Liu, Xue, Wang, Wu, Feng, Lu, Zhao, Deng, Zhang, Ruan, Dai, Chen, Ji, Li, Lin, Dai, Luo, Hao, Chen, Li, Zhang, Bao, Xu, Wang, Ding, Xin, Gao, Qu, Li, Guo, Li, Wang, Chen, Yuan, Qiu, Li, Cai, Ni, Liang, Chen, Dong, Hu, Gao, Guan, Huang, Yu, Wang, Zhang, Zhao, Wang, Zhang, Xu, Xia, Zhang, Zhang, Tang, Li, Wang, Li, Tian, Huang, Zhang, Wang, Chen, Du, Ge, Zhang, Pan, Wang, Chen, Jin, Chen, Lu, Zhou, Chen, Ye, Wang, Yu, Zhou, Pan, Li, Zhou, Wu, Ye, Yun, Pei, Sun, Wang, Zeng, Zhao, Liu, Liang, Gao, Yu, Zhang, Xiao, An, Liu, Wang, Chen, Nie, Cheng, Liu, Xie, Liu, Yang, Li, Su, Lin, Li, Jin, Shen, Chen, Sun, Wang, Song, Zhou, Wang, Shan, Li, Wang, Wei, Zhang, Xu, Li, Zhao, Sun, Wang, Yu, Zhang, Shi, Xiong, He, Piao, Wang, Tan, Ma, Liu, Guo, Ou, Wang, Gong, Zou, He, Xiong, Luo, You, Liu, Zhou, Zhu, Xu, Huang, Li, Zheng, Zhu, Ma, Tang, Zha, Yan, Ren, Ren, Sha, Fu, Xu, Xie, Zhang,
  Hao, Ma, Yan, Wu, Gu, Zhu, Liu, Li, Xie, Song, Pan, Huang, Xu, Zhang, and Zhang]{deepseekai2025deepseekr1incentivizingreasoningcapability}
DeepSeek-AI, Daya Guo, Dejian Yang, Haowei Zhang, Junxiao Song, Ruoyu Zhang, Runxin Xu, Qihao Zhu, Shirong Ma, Peiyi Wang, Xiao Bi, Xiaokang Zhang, Xingkai Yu, Yu~Wu, Z.~F. Wu, Zhibin Gou, Zhihong Shao, Zhuoshu Li, Ziyi Gao, Aixin Liu, Bing Xue, Bingxuan Wang, Bochao Wu, Bei Feng, Chengda Lu, Chenggang Zhao, Chengqi Deng, Chenyu Zhang, Chong Ruan, Damai Dai, Deli Chen, Dongjie Ji, Erhang Li, Fangyun Lin, Fucong Dai, Fuli Luo, Guangbo Hao, Guanting Chen, Guowei Li, H.~Zhang, Han Bao, Hanwei Xu, Haocheng Wang, Honghui Ding, Huajian Xin, Huazuo Gao, Hui Qu, Hui Li, Jianzhong Guo, Jiashi Li, Jiawei Wang, Jingchang Chen, Jingyang Yuan, Junjie Qiu, Junlong Li, J.~L. Cai, Jiaqi Ni, Jian Liang, Jin Chen, Kai Dong, Kai Hu, Kaige Gao, Kang Guan, Kexin Huang, Kuai Yu, Lean Wang, Lecong Zhang, Liang Zhao, Litong Wang, Liyue Zhang, Lei Xu, Leyi Xia, Mingchuan Zhang, Minghua Zhang, Minghui Tang, Meng Li, Miaojun Wang, Mingming Li, Ning Tian, Panpan Huang, Peng Zhang, Qiancheng Wang, Qinyu Chen, Qiushi Du, Ruiqi Ge, Ruisong
  Zhang, Ruizhe Pan, Runji Wang, R.~J. Chen, R.~L. Jin, Ruyi Chen, Shanghao Lu, Shangyan Zhou, Shanhuang Chen, Shengfeng Ye, Shiyu Wang, Shuiping Yu, Shunfeng Zhou, Shuting Pan, S.~S. Li, Shuang Zhou, Shaoqing Wu, Shengfeng Ye, Tao Yun, Tian Pei, Tianyu Sun, T.~Wang, Wangding Zeng, Wanjia Zhao, Wen Liu, Wenfeng Liang, Wenjun Gao, Wenqin Yu, Wentao Zhang, W.~L. Xiao, Wei An, Xiaodong Liu, Xiaohan Wang, Xiaokang Chen, Xiaotao Nie, Xin Cheng, Xin Liu, Xin Xie, Xingchao Liu, Xinyu Yang, Xinyuan Li, Xuecheng Su, Xuheng Lin, X.~Q. Li, Xiangyue Jin, Xiaojin Shen, Xiaosha Chen, Xiaowen Sun, Xiaoxiang Wang, Xinnan Song, Xinyi Zhou, Xianzu Wang, Xinxia Shan, Y.~K. Li, Y.~Q. Wang, Y.~X. Wei, Yang Zhang, Yanhong Xu, Yao Li, Yao Zhao, Yaofeng Sun, Yaohui Wang, Yi~Yu, Yichao Zhang, Yifan Shi, Yiliang Xiong, Ying He, Yishi Piao, Yisong Wang, Yixuan Tan, Yiyang Ma, Yiyuan Liu, Yongqiang Guo, Yuan Ou, Yuduan Wang, Yue Gong, Yuheng Zou, Yujia He, Yunfan Xiong, Yuxiang Luo, Yuxiang You, Yuxuan Liu, Yuyang Zhou, Y.~X. Zhu,
  Yanhong Xu, Yanping Huang, Yaohui Li, Yi~Zheng, Yuchen Zhu, Yunxian Ma, Ying Tang, Yukun Zha, Yuting Yan, Z.~Z. Ren, Zehui Ren, Zhangli Sha, Zhe Fu, Zhean Xu, Zhenda Xie, Zhengyan Zhang, Zhewen Hao, Zhicheng Ma, Zhigang Yan, Zhiyu Wu, Zihui Gu, Zijia Zhu, Zijun Liu, Zilin Li, Ziwei Xie, Ziyang Song, Zizheng Pan, Zhen Huang, Zhipeng Xu, Zhongyu Zhang, and Zhen Zhang.
\newblock Deepseek-r1: Incentivizing reasoning capability in llms via reinforcement learning, 2025.
\newblock URL \url{https://arxiv.org/abs/2501.12948}.

\bibitem[Dziri et~al.(2023)Dziri, Lu, Sclar, Li, Jiang, Lin, West, Bhagavatula, Le~Bras, Hwang, Sanyal, Welleck, Ren, Ettinger, Harchaoui, and Choi]{faithandfate}
Nouha Dziri, Ximing Lu, Melanie Sclar, Xiang~Lorraine Li, Liwei Jiang, Bill~Yuchen Lin, Peter West, Chandra Bhagavatula, Ronan Le~Bras, Jena~D. Hwang, Soumya Sanyal, Sean Welleck, Xiang Ren, Allyson Ettinger, Zaid Harchaoui, and Yejin Choi.
\newblock Faith and fate: limits of transformers on compositionality.
\newblock In \emph{Proceedings of the 37th International Conference on Neural Information Processing Systems}, NIPS '23, Red Hook, NY, USA, 2023. Curran Associates Inc.

\bibitem[Fong et~al.(2019)Fong, Spivak, and Tuy{\'{e}}ras]{DBLP:conf/lics/FongST19}
Brendan Fong, David~I. Spivak, and R{\'{e}}my Tuy{\'{e}}ras.
\newblock Backprop as functor: {A} compositional perspective on supervised learning.
\newblock In \emph{34th Annual {ACM/IEEE} Symposium on Logic in Computer Science, {LICS} 2019, Vancouver, BC, Canada, June 24-27, 2019}, pages 1--13. {IEEE}, 2019.
\newblock \doi{10.1109/LICS.2019.8785665}.
\newblock URL \url{https://doi.org/10.1109/LICS.2019.8785665}.

\bibitem[Fritz(2020)]{Fritz_2020}
Tobias Fritz.
\newblock A synthetic approach to markov kernels, conditional independence and theorems on sufficient statistics.
\newblock \emph{Advances in Mathematics}, 370:\penalty0 107239, August 2020.
\newblock ISSN 0001-8708.
\newblock \doi{10.1016/j.aim.2020.107239}.
\newblock URL \url{http://dx.doi.org/10.1016/j.aim.2020.107239}.

\bibitem[Gavranović et~al.(2024)Gavranović, Lessard, Dudzik, von Glehn, Araújo, and Veličković]{gavranović2024positioncategoricaldeeplearning}
Bruno Gavranović, Paul Lessard, Andrew Dudzik, Tamara von Glehn, João G.~M. Araújo, and Petar Veličković.
\newblock Position: Categorical deep learning is an algebraic theory of all architectures, 2024.
\newblock URL \url{https://arxiv.org/abs/2402.15332}.

\bibitem[Goldblatt(2006)]{goldblatt:topos}
Robert Goldblatt.
\newblock \emph{Topoi: The Categorial Analysis of Logic}.
\newblock Dover Press, 2006.

\bibitem[Gu et~al.(2022)Gu, Goel, and R{\'{e}}]{DBLP:conf/iclr/GuGR22}
Albert Gu, Karan Goel, and Christopher R{\'{e}}.
\newblock Efficiently modeling long sequences with structured state spaces.
\newblock In \emph{The Tenth International Conference on Learning Representations, {ICLR} 2022, Virtual Event, April 25-29, 2022}. OpenReview.net, 2022.
\newblock URL \url{https://openreview.net/forum?id=uYLFoz1vlAC}.

\bibitem[Hahn(2020)]{DBLP:journals/tacl/Hahn20}
Michael Hahn.
\newblock Theoretical limitations of self-attention in neural sequence models.
\newblock \emph{Trans. Assoc. Comput. Linguistics}, 8:\penalty0 156--171, 2020.
\newblock \doi{10.1162/TACL\_A\_00306}.
\newblock URL \url{https://doi.org/10.1162/tacl\_a\_00306}.

\bibitem[Johnstone(2014)]{Johnstone:topostheory}
Peter~T Johnstone.
\newblock \emph{{Topos Theory}}.
\newblock Dover Publications, 2014.

\bibitem[Kushner and Yin(2003)]{kushner2003stochastic}
H.~Kushner and G.G. Yin.
\newblock \emph{Stochastic Approximation and Recursive Algorithms and Applications}.
\newblock Stochastic Modelling and Applied Probability. Springer New York, 2003.
\newblock ISBN 9780387008943.
\newblock URL \url{https://books.google.com/books?id=_0bIieuUJGkC}.

\bibitem[MacLane(1971)]{maclane:71}
Saunders MacLane.
\newblock \emph{Categories for the Working Mathematician}.
\newblock Springer-Verlag, New York, 1971.
\newblock Graduate Texts in Mathematics, Vol. 5.

\bibitem[MacLane and leke Moerdijk(1994)]{maclane:sheaves}
Saunders MacLane and leke Moerdijk.
\newblock \emph{Sheaves in Geometry and Logic: A First Introduction to Topos Theory}.
\newblock Springer, 1994.

\bibitem[Mahadevan(2024)]{mahadevan2024gaiacategoricalfoundationsgenerative}
Sridhar Mahadevan.
\newblock Gaia: Categorical foundations of generative ai, 2024.
\newblock URL \url{https://arxiv.org/abs/2402.18732}.

\bibitem[Merrill et~al.(2022{\natexlab{a}})Merrill, Sabharwal, and Smith]{merrill-etal-2022-saturated}
William Merrill, Ashish Sabharwal, and Noah~A. Smith.
\newblock Saturated transformers are constant-depth threshold circuits.
\newblock \emph{Transactions of the Association for Computational Linguistics}, 10:\penalty0 843--856, 2022{\natexlab{a}}.
\newblock \doi{10.1162/tacl_a_00493}.
\newblock URL \url{https://aclanthology.org/2022.tacl-1.49/}.

\bibitem[Merrill et~al.(2022{\natexlab{b}})Merrill, Sabharwal, and Smith]{merrill2022saturatedtransformersconstantdepththreshold}
William Merrill, Ashish Sabharwal, and Noah~A. Smith.
\newblock Saturated transformers are constant-depth threshold circuits, 2022{\natexlab{b}}.
\newblock URL \url{https://arxiv.org/abs/2106.16213}.

\bibitem[Morris et~al.(2023)Morris, Sohl-Dickstein, Fiedel, Warkentin, Dafoe, Faust, Farabet, and Legg]{agi-dm}
Meredith~Ringel Morris, Jascha Sohl-Dickstein, Noah Fiedel, Tris Warkentin, Allan Dafoe, Aleksandra Faust, Clement Farabet, and Shane Legg, editors.
\newblock \emph{Levels of AGI for Operationalizing Progress on the Path to AGI}, 2023.
\newblock Original arXiv title in November 2023 was "Levels of AGI": Operationalizing Progress on the Path to AGI. Final title for publication as a position paper at ICML 2024 is: Levels of AGI for Operataionalizing Progress on the Path to AGI.

\bibitem[P\'erez et~al.(2021)P\'erez, Barcel\'o, and Marinkovic]{attention-turing-complete}
Jorge P\'erez, Pablo Barcel\'o, and Javier Marinkovic.
\newblock Attention is turing-complete.
\newblock \emph{Journal of Machine Learning Research}, 22\penalty0 (75):\penalty0 1--35, 2021.
\newblock URL \url{http://jmlr.org/papers/v22/20-302.html}.

\bibitem[Ram et~al.(2024)Ram, Klinger, and Gray]{llmcompositionality}
Parikshit Ram, Tim Klinger, and Alexander~G. Gray.
\newblock What makes models compositional? a theoretical view.
\newblock In \emph{Proceedings of the Thirty-Third International Joint Conference on Artificial Intelligence}, IJCAI '24, 2024.
\newblock ISBN 978-1-956792-04-1.
\newblock \doi{10.24963/ijcai.2024/533}.
\newblock URL \url{https://doi.org/10.24963/ijcai.2024/533}.

\bibitem[Richter(2020)]{richter2020categories}
B.~Richter.
\newblock \emph{From Categories to Homotopy Theory}.
\newblock Cambridge Studies in Advanced Mathematics. Cambridge University Press, 2020.
\newblock ISBN 9781108479622.
\newblock URL \url{https://books.google.com/books?id=pnzUDwAAQBAJ}.

\bibitem[Riehl(2017)]{riehl2017category}
E.~Riehl.
\newblock \emph{Category Theory in Context}.
\newblock Aurora: Dover Modern Math Originals. Dover Publications, 2017.
\newblock ISBN 9780486820804.
\newblock URL \url{https://books.google.com/books?id=6B9MDgAAQBAJ}.

\bibitem[Shaw et~al.(2018)Shaw, Uszkoreit, and Vaswani]{shaw2018selfattentionrelativepositionrepresentations}
Peter Shaw, Jakob Uszkoreit, and Ashish Vaswani.
\newblock Self-attention with relative position representations, 2018.
\newblock URL \url{https://arxiv.org/abs/1803.02155}.

\bibitem[Vaswani et~al.(2017)Vaswani, Shazeer, Parmar, Uszkoreit, Jones, Gomez, Kaiser, and Polosukhin]{DBLP:conf/nips/VaswaniSPUJGKP17}
Ashish Vaswani, Noam Shazeer, Niki Parmar, Jakob Uszkoreit, Llion Jones, Aidan~N. Gomez, Lukasz Kaiser, and Illia Polosukhin.
\newblock Attention is all you need.
\newblock In Isabelle Guyon, Ulrike von Luxburg, Samy Bengio, Hanna~M. Wallach, Rob Fergus, S.~V.~N. Vishwanathan, and Roman Garnett, editors, \emph{Advances in Neural Information Processing Systems 30: Annual Conference on Neural Information Processing Systems 2017, December 4-9, 2017, Long Beach, CA, {USA}}, pages 5998--6008, 2017.
\newblock URL \url{https://proceedings.neurips.cc/paper/2017/hash/3f5ee243547dee91fbd053c1c4a845aa-Abstract.html}.

\bibitem[Wang et~al.(2024)Wang, Wang, Athiwaratkun, Zhang, and Zou]{wang2024mixtureofagentsenhanceslargelanguage}
Junlin Wang, Jue Wang, Ben Athiwaratkun, Ce~Zhang, and James Zou.
\newblock Mixture-of-agents enhances large language model capabilities, 2024.
\newblock URL \url{https://arxiv.org/abs/2406.04692}.

\bibitem[Yun et~al.(2020)Yun, Bhojanapalli, Rawat, Reddi, and Kumar]{DBLP:conf/iclr/YunBRRK20}
Chulhee Yun, Srinadh Bhojanapalli, Ankit~Singh Rawat, Sashank~J. Reddi, and Sanjiv Kumar.
\newblock Are transformers universal approximators of sequence-to-sequence functions?
\newblock In \emph{8th International Conference on Learning Representations, {ICLR} 2020, Addis Ababa, Ethiopia, April 26-30, 2020}. OpenReview.net, 2020.
\newblock URL \url{https://openreview.net/forum?id=ByxRM0Ntvr}.

\end{thebibliography}

\newpage 

\section{Appendix: Mathematical Background}

\label{introcat}

Category theory is based fundamentally on defining {\em universal properties} \citep{riehl2017category}, which can be defined as the {\em initial} or {\em final} object in some category. To take a simple example, the Cartesian product of two sets can be defined as the set of ordered pairs, which tells us what it is, but not what it is good for, or why it is special in some way. Alternatively, we can define the Cartesian product of two sets as an object in the category {\bf Sets} that has the unique property that every function onto those sets must decompose uniquely as a composition of a function into the Cartesian product object, and then a projection component onto each component set. Furthermore, among all such objects that share this property, the Cartesian product is the final object. 

\begin{figure}[t]
\begin{center}
\begin{tabular}{|c |c | } \hline 
{\bf Set theory } & {\bf Topos theory} \\ \hline 
 set & object \\ 
 subset & subobject \\
 truth values $\{0, 1 \} $ & subobject classifier $\Omega$ \\
power set $P(A) = 2^A$ & power object $P(A) = \Omega^A$ \\ \hline
bijection & isomorphims \\ 
injection & monic arrow \\
surjection & epic arrow \\ \hline
singleton set $\{ * \}$ & terminal object ${\bf 1}$ \\ 
empty set $\emptyset$ & initial object ${\bf 0}$ \\
elements of a set $X$ & morphism $f: {\bf 1} \rightarrow X$ \\
- & functors, natural  transformations \\ 
- & limits, colimits, adjunctions \\ \hline
\end{tabular}
\end{center}
\caption{Comparison of notions from set theory and topos theory.} 
\label{setvscategories}
\end{figure} 

\begin{figure}[h]
\centering
      % Give a unique label
\includegraphics[scale=.5]{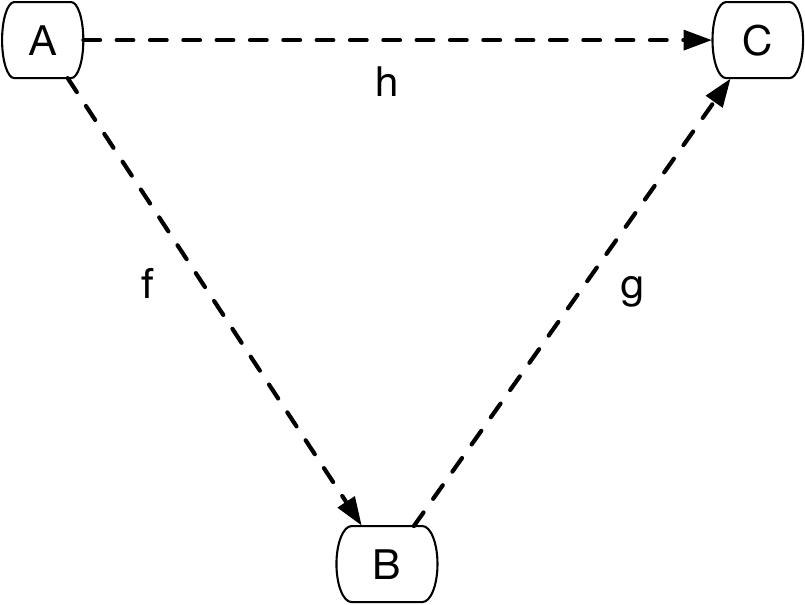}
\caption{Category theory is a compositional model of a system in terms of objects and their interactions.}
\label{morphisms} 
\end{figure}

Figure~\ref{setvscategories} compares the basic notions in set theory vs. category theory. Figure~\ref{morphisms} illustrates a simple category of 3 objects: $A$, $B$, and $C$ that interact through the morphisms $f: A \rightarrow B$, $g: B \rightarrow C$, and $h: A \rightarrow C$. Categories involve a fundamental notion of {\em composition}: the morphism $h: A \rightarrow C$ can be defined as the composition $g \circ f$ of the morphisms from $f$ and $g$. What the objects and morphisms represent is arbitrary, and like the canonical directed graph model, this abstractness gives category theory -- like graph theory -- a universal quality in terms of applicability to a wide range of problems. While categories and graphs and intimately related, in a category, there is no assumption of finiteness in terms of the cardinality of objects or morphisms. A category is defined to be {\em small} or {\em locally small} if there is a set's worth of objects and between any two objects, a set's worth of morphisms, but of course, a set need not be finite. As a simple example, the set of integers $\mathbb{Z}$ defines a category, where each integer $z$ is an object and there is a morphism $f: a \rightarrow b$ between integers $a$ and $b$ if $a \leq b$. This example serves to immediately clarify an important point: a category is only defined if both the objects and morphisms are defined. The category of integers $\mathbb{Z}$ may be defined in many ways, depending on what the morphisms represent. 

Briefly, a category is a collection of objects, and a collection of morphisms between pairs of objects, which are closed under composition, satisfy associativity, and include an identity morphism for every object. For example, sets form a category under the standard morphism of functions. Groups, modules, topological spaces and vector spaces all form categories in their own right, with suitable morphisms (e.g, for groups, we use group homomorphisms, and for vector spaces, we use linear transformations). 

A simple way to understand the definition of a category is to view it as a ``generalized" graph, where there is no limitation on the number of vertices, or the number of edges between any given pair of vertices. Each vertex defines an object in a category, and each edge is associated with a morphism. The underlying graph induces a ``free'' category where we consider all possible paths between pairs of vertices (including self-loops) as the set of morphisms between them. In the reverse direction, given a category, we can define a ``forgetful'' functor that extracts the underlying graph from the category, forgetting the composition rule. 

\begin{definition}
\label{cat-defn}
A {\bf {graph}} ${\cal G}$ (sometimes referred to as a quiver) is a labeled directed multi-graph defined by a set $O$ of {\em objects}, a set $A$ of {\em arrows},  along with two morphisms $s: A \rightarrow O$ and $t: A \rightarrow O$ that specify the domain and co-domain of each arrow.  In this graph, we define the set of composable pairs of arrows by the set 
\[ A \times_O A = \{\langle g, f  \rangle | \ g, f \in A, \ \ s(g) = t(f) \} \]

A {\bf {category}} ${\cal C}$ is a graph ${\cal G}$ with two additional functions: {${\bf id}:$} $O \rightarrow A$, mapping each object $c \in C$ to an arrow {${\bf id}_c$} and $\circ: A \times_O A \rightarrow A$, mapping each pair of composable morphisms $\langle f, g \rangle$ to their composition $g \circ f$. 
\end{definition}

It is worth emphasizing that no assumption is made here of the finiteness of a graph, either in terms of its associated objects (vertices) or arrows (edges). Indeed, it is entirely reasonable to define categories whose graphs contain an infinite number of edges. A simple example is the group $\mathbb{Z}$ of integers under addition, which can be represented as a single object, denoted $\{ \bullet \}$ and an infinite number of morphisms $f: \bullet \rightarrow \bullet$, each of which represents an integer, where composition of morphisms is defined by addition. In this example, all morphisms are invertible. In a general category with more than one object, a {\em groupoid} defines a category all of whose morphisms are invertible. A central principle in category theory is to avoid the use of equality, which is pervasive in mathematics, in favor of a more general notion of {\em isomorphism} or weaker versions of it. Many examples of categories can be given that are relevant to specific problems in AI and ML.  Some examples of categories of common mathematical structures are illustrated below. 

\begin{itemize} 

\item {\bf Set}: The canonical example of a category is {\bf Set}, which has as its objects, sets, and morphisms are functions from one set to another. The {\bf Set} category will play a central role in our framework, as it is fundamental to the universal representation constructed by Yoneda embeddings. 

\item {\bf Top:} The category {\bf Top} has topological spaces as its objects, and continuous functions as its morphisms. Recall that a topological space $(X, \Xi)$ consists of a set $X$, and a collection of subsets $\Xi$ of $X$ closed under finite intersection and arbitrary unions. 

\item {\bf Group:} The category {\bf Group} has groups as its objects, and group homomorphisms as its morphisms. 

\item {\bf Graph:} The category {\bf Graph} has graphs (undirected) as its objects, and graph morphisms (mapping vertices to vertices, preserving adjacency properties) as its morphisms. The category {\bf DirGraph} has directed graphs as its objects, and the morphisms must now preserve adjacency as defined by a directed edge. 

\item {\bf Poset:} The category {\bf Poset} has partially ordered sets as its objects and order-preserving functions as its morphisms. 

\item {\bf Meas:} The category {\bf Meas} has measurable spaces as its objects and measurable functions as its morphisms. Recall that a measurable space $(\Omega, {\cal B})$ is defined by a set $\Omega$ and an associated $\sigma$-field of subsets {\cal B} that is closed under complementation, and arbitrary unions and intersections, where the empty set $\emptyset \in {\cal B}$. 

\end{itemize}

Functors can be viewed as a generalization of the notion of morphisms across algebraic structures, such as groups, vector spaces, and graphs. Functors do more than functions: they not only map objects to objects, but like graph homomorphisms, they need to also map each morphism in the domain category to a corresponding morphism in the co-domain category. Functors come in two varieties, as defined below. 

 \begin{definition} 
A {\bf {covariant functor}} $F: {\cal C} \rightarrow {\cal D}$ from category ${\cal C}$ to category ${\cal D}$, and defined as \mbox{the following: }

\begin{itemize} 
    \item An object $F X$ (also written as $F(X)$) of the category ${\cal D}$ for each object $X$ in category ${\cal C}$.
    \item An  arrow  $F(f): F X \rightarrow F Y$ in category ${\cal D}$ for every arrow  $f: X \rightarrow Y$ in category ${\cal C}$. 
   \item The preservation of identity and composition: $F \ id_X = id_{F X}$ and $(F f) (F g) = F(g \circ f)$ for any composable arrows $f: X \rightarrow Y, g: Y \rightarrow Z$. 
\end{itemize}
\end{definition} 

\begin{definition} 
A {\bf {contravariant functor}} $F: {\cal C} \rightarrow {\cal D}$ from category ${\cal C}$ to category ${\cal D}$ is defined exactly like the covariant functor, except all the arrows are reversed. 
\end{definition} 

The {\em functoriality} axioms dictate how functors have to be behave: 

\begin{itemize} 

\item For any composable pair $f, g$ in category $C$, $Fg \cdot Ff = F(g \cdot f) $.

\item For each object $c$ in $C$, $F (1_c) = 1_{Fc}$.

\end{itemize}

\subsection{Natural Transformations and Universal Arrows}

Given any two functors $F: C \rightarrow D$ and $G: C \rightarrow D$ between the same pair of categories, we can define a mapping between $F$ and $G$ that is referred to as a natural transformation. These are defined through a collection of mappings, one for each object $c$ of $C$, thereby defining a morphism in $D$ for each object in $C$. 

\begin{definition}
    Given categories $C$ and $D$, and functors $F, G: C \rightarrow D$, a {\bf natural transformation} $\alpha: F \Rightarrow G$ is defined by the following data: 

    \begin{itemize}
        \item an arrow $\alpha_c: Fc \rightarrow Gc$ in $D$ for each object $c \in C$, which together define the components of the natural transformation. 
        \item For each morphism $f: c \rightarrow c'$, the following commutative diagram holds true: 

        % https://q.uiver.app/#q=WzAsNCxbMCwwLCJGYyJdLFszLDAsIkdjIl0sWzAsMiwiRmMnIl0sWzMsMiwiR2MnIl0sWzAsMSwiXFxhbHBoYV9jIl0sWzAsMiwiRmYiLDJdLFsyLDMsIlxcYWxwaGFfe2MnfSIsMl0sWzEsMywiR2YiXV0=
\[\begin{tikzcd}
	Fc &&& Gc \\
	\\
	{Fc'} &&& {Gc'}
	\arrow["{\alpha_c}", from=1-1, to=1-4]
	\arrow["Ff"', from=1-1, to=3-1]
	\arrow["{\alpha_{c'}}"', from=3-1, to=3-4]
	\arrow["Gf", from=1-4, to=3-4]
\end{tikzcd}\]

    \end{itemize}
    A {\bf natural isomorphism} is a natural transformation $\alpha: F \Rightarrow G$ in which every component $\alpha_c$ is an isomorphism. 
\end{definition}

This process of going from a category to its underlying directed graph  embodies a fundamental universal construction in category theory, called the {\em {universal arrow}}. It lies at the heart of many useful results, principally the Yoneda lemma that shows how object identity itself emerges from the structure of morphisms that lead into (or out of) it. 

\begin{definition}
Given a functor $S: D \rightarrow C$ between two categories, and an object $c$ of category $C$, a {\bf {universal arrow}} from $c$ to $S$ is a pair $\langle r, u \rangle$, where $r$ is an object of $D$ and $u: c \rightarrow Sr$ is an arrow of $C$, such that the following universal property holds true: 

\begin{itemize} 
\item For every pair $\langle d, f \rangle$ with $d$ an object of $D$ and $f: c \rightarrow Sd$ an arrow of $C$, there is a unique arrow $f': r \rightarrow d$ of $D$ with $S f' \circ u = f$. 
\end{itemize}
\end{definition}

\begin{definition}
If $D$ is a category and $H: D \rightarrow$ {\bf {Set}} is a set-valued functor, a {\bf {universal element}} associated with the functor $H$ is a pair $\langle r, e \rangle$ consisting of an object $r \in D$ and an element $e \in H r$ such that for every pair $\langle d, x \rangle$ with $x \in H d$, there is a unique arrow $f: r \rightarrow d$ of $D$ such \mbox{that $(H f) e = x$. }
\end{definition}

\begin{example}
Let $E$ be an equivalence relation on a set $S$, and consider the quotient set $S/E$ of equivalence classes, where $p: S \rightarrow S/E$ sends each element $s \in S$ into its corresponding equivalence class. The set of equivalence classes $S/E$ has the property that any function $f: S \rightarrow X$ that respects the equivalence relation can be written as $f s = f s'$ whenever $s \sim_E s'$, that is, $f = f' \circ p$, where the unique function $f': S/E \rightarrow X$. Thus, $\langle S/E, p \rangle$ is a universal element for the functor $H$. 
\end{example}

\begin{figure}[t] 
\centering
\begin{minipage}{0.5\textwidth}
\includegraphics[scale=0.3]{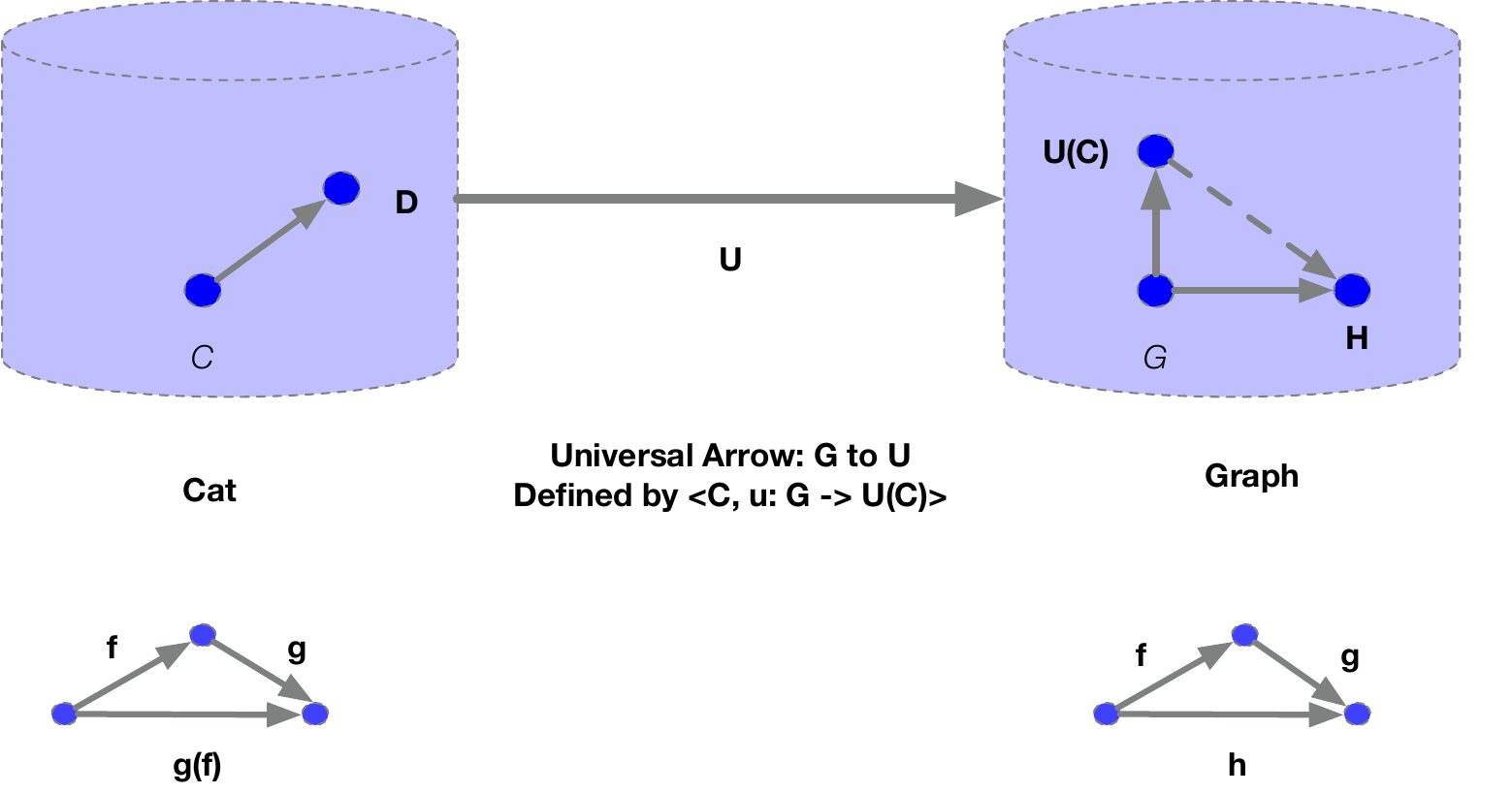}
\end{minipage}
\caption{The concept of universal arrows is illustrated through the connection between directed graphs, and their associated ``free" categories. In this example, the forgetful functor $U$ between {\bf {Cat}}, the category of all categories, and {\bf {Graph}}, the category of all (directed) graphs, maps any category into its underlying graph, forgetting which arrows are primitive and which are compositional.  The universal arrow from a graph $G$ to the forgetful functor $U$  is defined as a pair $\langle C, u: G \rightarrow U(C) \rangle$, where $u$ is a ``universal''  graph homomorphism. The universal arrow property asserts that every graph homomorphism $\phi: G \rightarrow H$ uniquely factors through the universal graph homomorphism $u: G \rightarrow U(C)$, where $U(C)$ is the graph induced  by category $C$ defining the universal arrow property. In other words, the associated {\em {extension}} problem of ``completing'' the triangle of graph homomorphisms in the category of {\bf {Graph}} can be uniquely solved by ``lifting'' the associated category arrow $h: C \rightarrow D$.}
\label{universal-arrow}
 \end{figure}

Figure~\ref{universal-arrow} illustrates the concept of universal arrows through the connection between categories and graphs. For every (directed) graph $G$, there is a universal arrow from $G$ to the ``forgetful'' functor $U$ mapping the category {\bf {Cat}} of all categories to {\bf {Graph}}, the category of all (directed) graphs, where for any category $C$, its associated graph is defined by $U(C)$. Intuitively, this forgetful functor ``throws'' away all categorical information, obliterating for example the distinction between the primitive morphisms $f$ and $g$ vs. their compositions $g \circ f$, both of which are simply viewed as edges in the graph $U(C)$.  To understand this functor,  consider a directed graph $U(C)$ defined from a category $C$, forgetting the rule for composition. That is, from the category $C$, which associates to each pair of composable arrows $f$ and $g$, the composed arrow $g \circ f$, we derive the underlying graph $U(C)$ simply by forgetting which edges correspond to elementary arrows, such as $f$ or $g$, and which are composites. For example, consider a partial order as the category ${\cal C}$, and then define  $U({\cal C} )$ as the directed graph that results from the transitive closure of the partial ordering.

 The universal arrow from a graph $G$ to the forgetful functor $U$  is defined as a pair $\langle C, u: G \rightarrow U(C) \rangle$, where $u$ is a ``universal''  graph homomorphism. This arrow possesses the following {\em {universal property}}: for every other pair $\langle D, v: G \rightarrow H \rangle$, where $D$ is a category, and $v$ is an arbitrary graph homomorphism, there is a functor  $f': C \rightarrow D$, which is an arrow in the category {\bf {Cat}} of all categories, such that {\em {every}} graph homomorphism $\phi: G \rightarrow H$ uniquely factors through the universal graph homomorphism $u: G \rightarrow U(C)$  as the solution to the equation $\phi = U(f') \circ u$, where $U(f'): U(C) \rightarrow H$ (that is, $H = U(D)$).  Namely, the dotted arrow defines a graph homomorphism $U(f')$ that makes the triangle diagram ``commute'', and the associated ``extension'' problem of finding this new graph homomorphism $U(f')$ is solved by ``lifting'' the associated category arrow $f': C \rightarrow D$. This property of universal arrows, as we show in the paper, provide the conceptual underpinnings of universal properties in many applications in AI and ML, as we will see throughout this paper.  

\subsection{Yoneda lemma and the Universality of Diagrams} 

The Yoneda Lemma  is one of the most celebrated results in category theory, and it provides a concrete example of the power of categorical thinking. Stated in simple terms, it states the mathematical objects are determined (up to isomorphism) by the interactions they make with other objects in a category. We will show the surprising results of applying this lemma to problems involving computing distances between objects in a metric space, reasoning about causal inference, and many other problems of importance in AI and ML. An analogy from particle physics proposed by Theo Johnson-Freyd might help give insight into this remarkable result: ``You work at a particle accelerator. You want to understand some particle. All you can do is throw other particles at it and see what happens. If you understand how your mystery particle responds to all possible test particles at all possible test energies, then you know everything there is to know about your mystery particle". The Yoneda Lemma states that the set of all morphisms into an object $d$ in a category ${\cal C}$, denoted as {\bf Hom}$_{\cal C}(-,d)$ and called the {\em contravariant functor} (or presheaf),  is sufficient to define $d$ up to isomorphism. The category of all presheaves forms a {\em category of functors}, and is denoted $\hat{{\cal C}} = $ {\bf Set}$^{{\cal C}^{op}}$.We will briefly describe two concrete applications of this lemma to two important areas in AI and ML in this section: reasoning about causality and reasoning about distances. The Yoneda lemma plays a crucial role in this paper because it defines the concept of a {\em universal representation} in category theory. We first show that associated with universal arrows is the corresponding induced isomorphisms between {\bf {Hom}} sets of morphisms in categories. This universal property then leads to the Yoneda lemma. 

\begin{theorem}
Given any functor $S: D \rightarrow C$, the universal arrow $\langle r, u: c \rightarrow Sr \rangle$ implies a bijection exists between the {\bf {Hom}} sets 
\[ \mbox{{\bf Hom}}_{D}(r, d) \simeq \mbox{{\bf Hom}}_{C}(c, Sd) \]
\end{theorem}

A special case of this natural transformation that transforms the identity morphism {\bf {1}}$_r$ leads us to the Yoneda lemma.
\begin{center}
 \begin{tikzcd}
  D(r,r) \arrow{d}{D(r, f')} \arrow{r}{\phi_r}
    & C(c, Sr) \arrow[]{d}{C(c, S f')} \\
  D(r,d)  \arrow[]{r}[]{\phi_d}
&C(c, Sd)\end{tikzcd}
 \end{center} 

 As the two paths shown here must be equal in a commutative diagram, we get the property that a bijection between the {\bf {Hom}} sets holds precisely when $\langle r, u: c \rightarrow Sr \rangle$ is a universal arrow from $c$ to $S$. Note that for the case when the categories $C$ and $D$ are small, meaning their {\bf Hom} collection of arrows forms a set, the induced functor {\bf {Hom}}$_C(c, S - )$ to {\bf Set} is isomorphic to the functor {\bf {Hom}}$_D(r, -)$. This type of isomorphism defines a universal representation, and is at the heart of the causal reproducing property (CRP) defined below. 

\begin{lemma}
{\bf {Yoneda lemma}}: For any functor $F: C \rightarrow {\bf Set}$, whose domain category $C$ is ``locally small" (meaning that the collection of morphisms between each pair of objects forms a set), any object $c$ in $C$, there is a bijection 

\[ \mbox{Hom}(C(c, -), F) \simeq Fc \]

that defines a natural transformation $\alpha: C(c, -) \Rightarrow F$ to the element $\alpha_c(1_c) \in Fc$. This correspondence is natural in both $c$ and $F$. 
\end{lemma}

There is of course a dual form of the Yoneda Lemma in terms of the contravariant functor $C(-, c)$ as well using the natural transformation $C(-, c) \Rightarrow F$. A very useful way to interpret the Yoneda Lemma is through the notion of universal representability through a covariant or contravariant functor.

\begin{definition}
    A {\bf universal representation} of an object $c \in C$ in a category $C$ is defined as a contravariant functor $F$ together with a functorial representation $C(-, c) \simeq F$ or by a covariant functor $F$ together with a representation $C(c, -) \simeq F$. The collection of morphisms $C(-, c)$ into an object $c$ is called the {\bf presheaf}, and from the Yoneda Lemma, forms a universal representation of the object. 
\end{definition}

Another useful concept was introduced by the mathematician Grothendieck, who made many important contributions to category theory. 

\begin{definition}
    The {\bf category of elements} $\int F$ of a covariant functor $F: C \rightarrow \mbox{{\bf Set}}$ is defined as

    \begin{itemize}
        \item a collection of objects  $(c, x)$ where $c \in C$ and $x \in Fc$

        \item a collection of morphisms $(c, x) \rightarrow (c', x')$ for every morphism $f: c \rightarrow c'$ such that $F f(x) = x'$. 
    \end{itemize}
\end{definition}

\begin{definition}
    The {\bf category of elements} $\int F$ of a contravariant functor $F: C^{op} \rightarrow \mbox{{\bf Set}}$ is defined as

    \begin{itemize}
        \item a collection of objects  $(c, x)$ where $c \in C$ and $x \in Fc$

        \item a collection of morphisms $(c, x) \rightarrow (c', x')$ for every morphism $f: c \rightarrow c'$ such that $F f(x') = x$. 
    \end{itemize}
\end{definition}

There is a natural ``forgetful" functor $\pi: \int F \rightarrow C$ that maps the pairs of objects $(c, x) \in \int F$ to $c \in C$ and maps morphisms $(c, x) \rightarrow (c', x') \in \int F$ to $f: c \rightarrow c' \in C$. Below we will show that the category of elements $\int F$ can be defined through a universal construction as the pullback in the diagram of categories.

A key distinguishing feature of category theory is the use of diagrammatic reasoning. However, diagrams are also viewed more abstractly as functors mapping from some indexing category to the actual category. Diagrams are useful in understanding universal constructions, such as limits and colimits of diagrams. To make this somewhat abstract definition concrete, let us look at some simpler examples of universal properties, including co-products and quotients (which in set theory correspond to disjoint unions). Coproducts refer to the universal property of abstracting a group of elements into a larger one.

 Before we formally the concept of limit and colimits, we consider some examples.  These notions generalize the more familiar notions of Cartesian products and disjoint unions in the category of {\bf {Sets}}, the notion of meets and joins in the category {\bf {Preord}} of preorders, as well as the  least upper bounds and greatest lower bounds in lattices, and many other concrete examples from mathematics. 

\begin{example} 
If  we consider a small  ``discrete'' category ${\cal D}$ whose only morphisms are identity arrows, then the colimit of a functor ${\cal F}: {\cal D} \rightarrow {\cal C}$ is the {\em categorical coproduct} of ${\cal F}(D)$ for $D$, an object of category {\cal D}, is denoted as 
\[ \mbox{Colimit}_{\cal D} F = \bigsqcup_D {\cal F}(D) \]

In the special case when the category {\cal C} is the category {\bf {Sets}}, then the colimit of this functor is simply the disjoint union of all the sets $F(D)$ that are mapped from objects $D \in {\cal D}$. 
\end{example} 

\begin{example} 
Dual to the notion of colimit of a functor is the notion of {\em limit}. Once again, if we consider a small  ``discrete'' category ${\cal D}$ whose only morphisms are identity arrows, then the limit of a functor ${\cal F}: {\cal D} \rightarrow {\cal C}$ is the {\em categorical product} of ${\cal F}(D)$ for $D$, an object of category {\cal D}, is denoted as 
\[ \mbox{limit}_{\cal D} F = \prod_D {\cal F}(D) \]

In the special case when the category {\cal C} is the category {\bf {Sets}}, then the limit of this functor is simply the Cartesian product of all the sets $F(D)$ that are mapped from objects $D \in {\cal D}$. 
\end{example} 

Category theory relies extensively on {\em universal constructions}, which satisfy a universal property. One of the central building blocks is the identification of universal properties through formal diagrams.  Before introducing these definitions in their most abstract form, it greatly helps to see some simple examples. 

 We can illustrate the limits and colimits in diagrams using pullback and pushforward mappings.

\begin{tikzcd}
%  T
%  \arrow[drr, bend left, "x"]
%  \arrow[ddr, bend right, "y"]
%  \arrow[dr, dotted, "r" description] & & \\
    & Z\arrow[r, "p"] \arrow[d, "q"]
      & X \arrow[d, "f"] \arrow[ddr, bend left, "h"]\\
& Y \arrow[r, "g"] \arrow[drr, bend right, "i"] &X \sqcup Y \arrow[dr, "r"]  \\ 
& & & R 
\end{tikzcd}

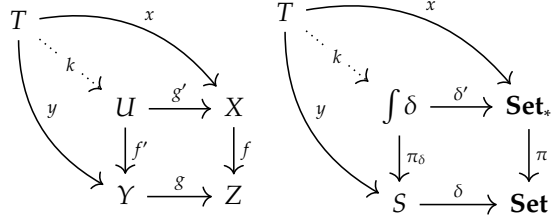
\begin{figure}[h]
\centering
\begin{tikzcd}
  T
  \arrow[drr, bend left, "x"]
  \arrow[ddr, bend right, "y"]
  \arrow[dr, dotted, "k" description] & & \\
    & 
    U\arrow[r, "g'"] \arrow[d, "f'"]
      & X \arrow[d, "f"] \\
& Y \arrow[r, "g"] &Z
\end{tikzcd}
\begin{tikzcd}
  T
  \arrow[drr, bend left, "x"]
  \arrow[ddr, bend right, "y"]
  \arrow[dr, dotted, "k" description] & & \\
    & 
    \int \delta \arrow[r, "\delta'"] \arrow[d, "\pi_\delta"]
      & {\bf Set}_* \arrow[d, "\pi"] \\
& S \arrow[r, "\delta"] & {\bf Set}
\end{tikzcd}
\caption{(\textbf{Left})
 Universal Property of pullback mappings. (\textbf{Right}) The Grothendieck category of elements $\int \delta$ of any set-valued functor $\delta: S \rightarrow$ {\bf {Set}} can be described as a pullback in the diagram of categories. Here, {\bf Set}$_*$ is the category of pointed sets $(X, x \in X)$, and $\pi$ is the ``forgetful" functor that sends a pointed set $(X, x \in X)$ into the underlying set $X$.  } 
\label{univpr}
\end{figure}

An example of a universal construction is given by the above commutative diagram, where the coproduct object $X \sqcup Y$ uniquely factorizes any mapping $h: X \rightarrow R$, such that any mapping $i: Y \rightarrow R$, so that $h = r \circ f$, and furthermore $i = r \circ g$. Co-products are themselves special cases of the more general notion of co-limits. Figure~\ref{univpr}  illustrates the fundamental property of a {\em {pullback}}, which along with {\em pushforward}, is one of the core ideas in category theory. The pullback square with the objects $U,X, Y$ and $Z$ implies that the composite mappings $g \circ f'$ must equal $g' \circ f$. In this example, the morphisms $f$ and $g$ represent a {\em {pullback}} pair, as they share a common co-domain $Z$. The pair of morphisms $f', g'$ emanating from $U$ define a {\em {cone}}, because the pullback square ``commutes'' appropriately. Thus, the pullback of the pair of morphisms $f, g$ with the common co-domain $Z$ is the pair of morphisms $f', g'$ with common domain $U$. Furthermore, to satisfy the universal property, given another pair of morphisms $x, y$ with common domain $T$, there must exist another morphism $k: T \rightarrow U$ that ``factorizes'' $x, y$ appropriately, so that the composite morphisms $f' \ k = y$ and $g' \ k = x$. Here, $T$ and $U$ are referred to as {\em cones}, where $U$ is the limit of the set of all cones ``above'' $Z$. If we reverse arrow directions appropriately, we get the corresponding notion of pushforward. So, in this example, the pair of morphisms $f', g'$ that share a common domain represent a pushforward pair. 
As Figure~\ref{univpr}, for any set-valued functor $\delta: S \rightarrow$ {\bf {Sets}}, the Grothendieck category of elements $\int \delta$ can be shown to be a pullback in the diagram of categories. Here, {${\bf Set}_*$} is the category of pointed sets, and $\pi$ is a projection that sends a pointed set $(X, x \in X)$ to its \mbox{underlying set $X$.}

We can now proceed to define limits and colimits more generally. We define a {\em diagram} $F$ of {\em shape} $J$ in a category $C$ formally as a functor $F: J \rightarrow C$. We want to define the somewhat abstract concepts of {\em limits} and {\em colimits}, which will play a central role in this paper in identifying properties of AI and ML techniques.  A convenient way to introduce these concepts is through the use of {\em universal cones} that are {\em over} and {\em under} a diagram.

For any object $c \in C$ and any category $J$, the {\em constant functor} $c: J \rightarrow C$ maps every object $j$ of $J$ to $c$ and every morphism $f$ in $J$ to the identity morphisms $1_c$. We can define a constant functor embedding as the collection of constant functors $\Delta: C \rightarrow C^J$ that send each object $c$ in $C$ to the constant functor at $c$ and each morphism $f: c \rightarrow c'$ to the constant natural transformation, that is, the natural transformation whose every component is defined to be the morphism $f$. 

Cones under a diagram are referred to usually as {\em cocones}. Using the concept of cones and cocones, we can now formally define the concept of limits and colimits more precisely. 

\begin{definition}
    For any diagram $F: J \rightarrow C$, there is a functor 

    \[ \mbox{Cone}(-, F): C^{op} \rightarrow \mbox{{\bf Set}} \]

    which sends $c \in C$ to the set of cones over $F$ with apex $c$. Using the Yoneda Lemma, a {\bf limit} of $F$ is defined as an object $\lim F \in C$ together with a natural transformation $\lambda: \lim F \rightarrow F$, which can be called the {\bf universal cone} defining the natural isomorphism 

    \[ C(-, \lim F) \simeq \mbox{Cone}(-, F) \]

    Dually, for colimits, we can define a functor 

    \[ \mbox{Cone}(F, -): C \rightarrow \mbox{{\bf Set}} \]

    that maps object $c \in C$ to the set of cones under $F$ with nadir $c$. A {\bf colimit} of $F$ is a representation for $\mbox{Cone}(F, -)$. Once again, using the Yoneda Lemma, a colimit is defined by an object $\mbox{Colim} F \in C$ together with a natural transformation $\lambda: F \rightarrow \mbox{colim} F$, which defines the {\bf colimit cone} as the natural isomorphism 

    \[ C(\mbox{colim} F, -) \simeq \mbox{Cone}(F, -) \]
\end{definition}

Limit and colimits of diagrams over arbitrary categories can often be reduced to the case of their corresponding diagram properties over sets. One important stepping stone is to understand how functors interact with limits and colimits. 

\begin{definition}
    For any class of diagrams $K: J \rightarrow C$, a functor $F: C \rightarrow D$ 

    \begin{itemize}
        \item {\bf preserves} limits if for any diagram $K: J \rightarrow C$ and limit cone over $K$, the image of the cone defines a limit cone over the composite diagram $F K: J \rightarrow D$. 

        \item {\bf reflects} limits if for any cone over a diagram $K: J \rightarrow C$ whose image upon applying $F$ is a limit cone for the diagram $F K: J \rightarrow D$ is a limit cone over $K$

        \item {\bf creates} limits if whenever $FK : J \rightarrow D$ has a limit in $D$, there is some limit cone over $F K$ that can be lifted to a limit cone over $K$ and moreoever $F$ reflects the limits in the class of diagrams. 
    \end{itemize}
\end{definition}

To interpret these abstract definitions, it helps to concretize them in terms of a specific universal construction, like the pullback defined above $c' \rightarrow c \leftarrow c''$ in $C$. Specifically, for pullbacks: 

\begin{itemize} 

\item A functor $F$ {\bf preserves pullbacks} if whenever $p$ is the pullback of  $c' \rightarrow c \leftarrow c''$ in $C$, it follows that $Fp$ is the pullback of  $Fc' \rightarrow Fc \leftarrow Fc''$ in $D$.

\item A functor $F$ {\bf reflects  pullbacks}  if  $p$ is the pullback of  $c' \rightarrow c \leftarrow c''$ in $C$ whenever $Fp$ is the pullback of  $Fc' \rightarrow Fc \leftarrow Fc''$ in $D$.

\item A functor $F$ {\bf creates pullbacks} if there exists some $p$ that is the pullback of  $c' \rightarrow c \leftarrow c''$ in $C$ whenever there exists a $d$ such  that $d$ is the pullback of  $Fc' \rightarrow Fc \leftarrow Fc''$ in $F$.

\end{itemize} 

\subsection*{Universality of Diagrams}

In the category {\bf {Sets}}, we know that every object (i.e., a set) $X$ can be expressed as a coproduct (i.e., disjoint union)  of its elements $X \simeq \sqcup_{x \in X} \{ x \}$, where $x \in X$. Note that we can view each element $x \in X$ as a morphism $x: \{ * \} \rightarrow X$ from the one-point set to $X$. The categorical generalization of this result is called the {\em {density theorem}} in the theory of sheaves. First, we define the key concept of a {\em comma category}. 

\begin{definition}
Let $F: {\cal D} \rightarrow {\cal C}$ be a functor from category ${\cal D}$ to ${\cal C}$. The {\bf {comma category}} $F \downarrow {\cal C}$ is one whose objects are pairs $(D, f)$, where $D \in {\cal D}$ is an object of ${\cal D}$ and $f \in$ {\bf {Hom}}$_{\cal C}(F(D), C)$, where $C$ is an object of ${\cal C}$. Morphisms in the comma category $F \downarrow {\cal C}$ from $(D, f)$ to $(D', f')$, where $g: D \rightarrow D'$, such that $f' \circ F(g) = f$. We can depict this structure through the following commutative diagram: 
\begin{center} 
\begin{tikzcd}[column sep=small]
& F(D) \arrow{dl}[near start]{F(g)} \arrow{dr}{f} & \\
  F(D')\arrow{rr}{f'}&                         & C
\end{tikzcd}
\end{center} 
\end{definition} 

We first introduce the concept of a {\em {dense}} functor: 

\begin{definition}
Let {\cal D} be a small category, {\cal C} be an arbitrary category, and $F: {\cal D} \rightarrow {\cal D}$ be a functor. The functor $F$ is {\bf {dense}} if for all objects $C$ of ${\cal C}$, the natural transformation 
\[ \psi^C_F: F \circ U \rightarrow \Delta_C, \ \ (\psi^C_F)_{({\cal D}, f)} = f\]
is universal in the sense that it induces an isomorphism $\mbox{Colimit}_{F \downarrow C} F \circ U \simeq C$. Here, $U: F \downarrow C \rightarrow {\cal D}$ is the projection functor from the comma category $F \downarrow {\cal C}$, defined by $U(D, f) = D$. 

\end{definition} 

A fundamental consequence of the category of elements is that every object in the functor category of presheaves, namely contravariant functors from a category into the category of sets, is the colimit of a diagram of representable objects, via the Yoneda lemma. Notice this is a generalized form of the density notion from the category {\bf {Sets}}.

\begin{theorem}
\label{presheaf-theorem}
{\bf {Universality of Diagrams}}: In the functor category of presheaves {\bf {Set}}$^{{\cal C}^{op}}$, every object $P$ is the colimit of a diagram of representable objects, in a canonical way. 
\end{theorem}

\subsection{Heyting Algebras} 

Subobject classifiers in a general topos category, such as the LLM category ${\cal C}^\rightarrow_T$, are not Boolean, and require intuitionistic semantics. We define the Heyting algebra, which is useful in the analysis of such subobject classifiers. 

\begin{definition}
    A {\bf Heyting algebra} is a poset with all finite products and coproducts, which is Cartesian closed. That is, a Heyting algebra is a lattice, including bottom and top elements, denoted by ${\bf 0}$ and ${\bf 1}$, respectively, which associates to each pair of elements $x$ and $y$ an exponential $y^x$. The exponential is written $x \Rightarrow y$, and defined as an adjoint functor: 
    \[ z \leq (x \Rightarrow y) \ \ \mbox{if and only if} \ \ z \wedge x \leq y\]
\end{definition}
In other words, $x \Rightarrow y$ is a least upper bound for all those elements $z$ with $z \wedge x \leq y$.  As a concrete example, for a topological space $X$ the set of open sets ${\cal O}(X)$ is a Heyting algebra. The binary intersections and unions of open sets yield open sets. The empty set $\emptyset$ represents ${\bf 0}$ and the complete set $X$ represents ${\bf 1}$. Given any two open sets $U$ and $V$, the exponential object $U \Rightarrow W$ is defined as the union $\bigcup_i W_i$ of all open sets $W_i$ for which $W \cap U \subset V$. 

% https://q.uiver.app/#q=WzAsNCxbMywwLCJ4IFxcUmlnaHRhcnJvdyB5Il0sWzIsMSwieSJdLFsxLDIsInggXFx3ZWRnZSB5Il0sWzAsMSwieCJdLFsyLDFdLFsxLDBdLFsyLDNdXQ==
%\[\begin{tikzcd}
%	&&& {x \Rightarrow y} \\
%	x && y \\
%	& {x \wedge y}
%	\arrow[from=3-2, to=2-3]
%	\arrow[from=2-3, to=1-4]
%	\arrow[from=3-2, to=2-1]
%\end{tikzcd}\]

Note that in a Boolean algebra, we define implication as the relationship $(x \Rightarrow y) \equiv \neg x \vee y$. This property is referred to as the ``law of the excluded middle" (because if $x = y$, then this translates to $\neg x \vee x = {\bf true}$) does not always hold in a Heyting algebra. 
%As Figure~\ref{fig:gdc-sieves} illustrated, when GDC is defined using sieves over a category, causal interventions become subobjects in  a partial order induced by the underlying Grothendieck topology. 
%Even in the simple case of SCMs, we noted above that subobject classifiers have ``multiple degrees of truth", beyond Boolean algebra.  
%

\subsection{Monoidal Categories} 

We sketched out the implementation of topos-theoretic LLM architectures using the functorial framework of backpropagation introduced by \citet{DBLP:conf/lics/FongST19}. This framework uses symmetric monoidal categories, which we briefly review here. 

%SMCOMMENT: Boldface removed. 
\begin{definition}
    A monoidal category is a category {\cal C} together with a functor $\otimes: {\cal C} \times {\cal C} \rightarrow {\cal C}$, an identity object $e$ of {\cal C}, and natural isomorphisms $\alpha, \lambda, \rho$ defined as follows: 
    \begin{eqnarray*}
        \alpha_{C_1, C_2, C_3}: C_1 \otimes (C_2 \otimes C_3) & \cong & (C_1 \otimes C_2) \otimes C_3 \\
        \lambda_C: e \otimes C & \cong & C, \ \ \mbox{for all objects} \  \ C  \\
        \rho: C \otimes e & \cong & C, \ \  \mbox{for all objects} \ C
    \end{eqnarray*}
\end{definition}

The natural isomorphisms must satisfy coherence conditions called the ``pentagon'' and ``triangle'' diagrams~\citep{maclane:71}.
%SMCOMMENT: Boldface removed 
\begin{definition}
    A symmetric monoidal category is a monoidal category $({\cal C}, \otimes, e, \alpha, \lambda, \rho)$, together with a natural isomorphism 
    \begin{eqnarray*}
       \tau_{C_1, C_2}: C_1 \otimes C_2 \cong C_2 \otimes C_1, \ \ \mbox{for all objects} \ \ C_1, C_2
    \end{eqnarray*}
    where $\tau$ satisfies the additional conditions: for all objects $C_1, C_2$ $\tau_{C_2, C_1} \circ \tau_{C_1, C_2} \cong 1_{C_1 \otimes C_2}$ and for all objects $C$, $\rho_C = \lambda_C \circ \tau_{C, e}: C \otimes e \cong C$. 
\end{definition}

\end{document}